\documentclass[letterpaper]{article} 
\usepackage{aaai2026}  
\usepackage{times}  
\usepackage{helvet}  
\usepackage{courier}  
\usepackage[hyphens]{url}  
\usepackage{graphicx} 
\usepackage{natbib}  
\usepackage{caption} 
\usepackage{algorithm}
\usepackage{algorithmic}

\usepackage{newfloat}
\usepackage{listings}
\DeclareCaptionStyle{ruled}{labelfont=normalfont,labelsep=colon,strut=off} 
\floatstyle{ruled}
\newfloat{listing}{tb}{lst}{}
\floatname{listing}{Listing}
\usepackage{colortbl}
\usepackage{booktabs}       %
\usepackage{amsfonts}       %
\usepackage{nicefrac}       %
\usepackage{microtype}      %
\usepackage{minitoc}
\usepackage{pgfplots} 
\usepackage{amsmath}
\usepackage{subcaption}
\usepackage{enumitem} %
\usepackage{makecell}
\usepackage{tabularx}
\usepackage{adjustbox}
\usepackage{anyfontsize}
\usepackage{tikz}  %
\usepackage{rotating} %
\usepackage{amssymb}  %
\usepackage{multirow} %
\usepackage{array} %
\usepackage{lipsum} %
\usepackage{cleveref}
\usepackage{titletoc}
\usepackage[most]{tcolorbox}

\newenvironment{qbox}
{\begin{tcolorbox}[enhanced jigsaw, drop shadow=black!50!white,colback=white, width=0.95\linewidth, center, left=2pt,right=2pt,top=1pt,bottom=1pt]}
{\end{tcolorbox}}

\definecolor{colorred}{cmyk}{0, 0.652, 0.615, 0.031}

\definecolor{colorblue}{cmyk}{0.984, 0.424, 0, 0.243}

\definecolor{orangea}{cmyk}{0, 0.18, 0.305, 0.023}
\newcommand {\orangea}[1]{{\cellcolor{orangea}{#1}}}
\definecolor{orangeb}{cmyk}{0, 0.092, 0.164, 0.019}
\newcommand {\orangeb}[1]{{\cellcolor{orangeb}{#1}}}
\definecolor{orangec}{cmyk}{0, 0.064, 0.12, 0.016}
\newcommand {\orangec}[1]{{\cellcolor{orangec}{#1}}}
\definecolor{oranged}{cmyk}{0, 0.032, 0.059, 0.007}
\newcommand {\oranged}[1]{{\cellcolor{oranged}{#1}}}
\definecolor{orangee}{cmyk}{0, 0.024, 0.043, 0.007}
\newcommand {\orangee}[1]{{\cellcolor{orangee}{#1}}}
\definecolor{orangef}{cmyk}{0, 0, 0, 0}
\newcommand {\orangef}[1]{{\cellcolor{orangef}{#1}}}

\definecolor{bluea}{cmyk}{0.193, 0.088, 0, 0.066}
\newcommand {\bluea}[1]{{\cellcolor{bluea}{#1}}}
\definecolor{blueb}{cmyk}{0.159, 0.071, 0, 0.062}
\newcommand {\blueb}[1]{{\cellcolor{blueb}{#1}}}
\definecolor{bluec}{cmyk}{0.114, 0.053, 0, 0.039}
\newcommand {\bluec}[1]{{\cellcolor{bluec}{#1}}}
\definecolor{blued}{cmyk}{0.086, 0.043, 0, 0.039}
\newcommand {\blued}[1]{{\cellcolor{blued}{#1}}}
\definecolor{bluee}{cmyk}{0.024, 0.012, 0, 0.011}
\newcommand {\bluee}[1]{{\cellcolor{bluee}{#1}}}
\definecolor{bluef}{cmyk}{0, 0, 0, 0}
\newcommand {\bluef}[1]{{\cellcolor{bluef}{#1}}}
\newcommand{\DALLE}{{DALL$\cdot$E}}

\title{HiGFA: Hierarchical Guidance for Fine-grained Data \\ Augmentation with Diffusion Models}
\author {
    Zhiguang Lu\textsuperscript{\rm 1,\rm 2}, 
    Qianqian Xu\textsuperscript{\rm 1,\rm 5\thanks{Corresponding Authors}}, 
    Peisong Wen\textsuperscript{\rm 2}, 
    Siran Dai\textsuperscript{\rm 3,\rm 4}, 
    Qingming Huang\textsuperscript{\rm 2,\rm 1*}
}
\affiliations {
    \textsuperscript{\rm 1}State Key Laboratory of AI Safety, Institute of Computing Technology, Chinese Academy of Sciences \\
    \textsuperscript{\rm 2}School of Computer Science and Technology, University of Chinese Academy of Sciences \\
    \textsuperscript{\rm 3}Institute of Information Engineering, Chinese Academy of Sciences \\
    \textsuperscript{\rm 4}School of Cyber Security, University of Chinese Academy of Sciences \\
    \textsuperscript{\rm 5}Peng Cheng Laboratory \\
    \{luzhiguang23s, xuqianqian\}@ict.ac.cn, wenpeisong@ucas.ac.cn, daisiran@iie.ac.cn, qmhuang@ucas.ac.cn 
}

\usepackage{bibentry}

\begin{document}

\maketitle

\begin{abstract}
Generative diffusion models show promise for data augmentation. 
However, applying them to fine-grained tasks presents a significant challenge: ensuring synthetic images accurately capture the subtle, category-defining features critical for high fidelity.
Standard approaches, such as text-based Classifier-Free Guidance (CFG), often lack the required specificity, potentially generating misleading examples that degrade fine-grained classifier performance. 
To address this, we propose Hierarchically Guided Fine-grained Augmentation (HiGFA). 
HiGFA leverages the temporal dynamics of the diffusion sampling process. 
It employs strong text and transformed contour guidance with fixed strengths in the early-to-mid sampling stages to establish overall scene, style, and structure. 
In the final sampling stages, HiGFA activates a specialized fine-grained classifier guidance and dynamically modulates the strength of all guidance signals based on prediction confidence. 
This hierarchical, confidence-driven orchestration enables HiGFA to generate diverse yet faithful synthetic images by intelligently balancing global structure formation with precise detail refinement. 
Experiments on several FGVC datasets demonstrate the effectiveness of HiGFA.

\end{abstract}

\begin{links}
    \link{Code}{https://github.com/ZhiguangLuu/HiGFA}
\end{links}

\section{Introduction}
\label{sec:intro}


Deep learning models have achieved remarkable success across various domains, yet their performance in specialized tasks like fine-grained visual classification (FGVC) often hinges on large, diverse, and accurately labeled datasets. 
Acquiring such datasets can be prohibitively expensive and time-consuming. 
Data augmentation techniques are crucial for mitigating data scarcity and improving model generalization. 
While traditional augmentation methods, such as geometric transforms, color jittering, offer limited diversity, generative models, particularly diffusion models~\cite{ho2020denoising, song2021scorebased,rombach2022high, podell2024sdxl}, provide a promising alternative to synthesizing novel and realistic data samples. 
Typically, existing methods utilize class text or image structure as conditions to guide generation and assume the output matches the target category~\cite{trabucco2023effective, he2022synthetic,zhang2023addingcontrol, michaeli2024advancing}. 
Compared to traditional data augmentation, this technical route significantly increases data diversity, thereby improving model generalization in downstream tasks.

\begin{figure}[t]
    \centering
    \includegraphics[width=0.99\columnwidth]{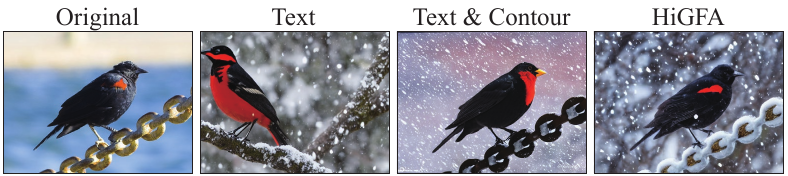}
    \caption{
        Accurately depicting the red shoulder patches of the Red-winged Blackbird is essential for fine-grained classification. More examples are provided in Appendix.
        }
        \label{fig:intro_vis}
\end{figure}

Unfortunately, without careful design, these methods are inferior to traditional augmentation in fine-grained visual classification tasks due to the \textbf{insufficient fidelity to fine-grained categories}. 
Specifically, fine-grained categories often differ only in subtle characteristics, such as the specific shape of a bird's beak, the precise pattern on an insect's wing, or subtle variations in car model designs. 
However, text descriptions and image contours fail to faithfully capture these subtle and category-defining visual attributes. 
For instance, as illustrated in \Cref{fig:intro_vis}, generating an image labeled ``Red-winged Blackbird'' that omits its distinctive red shoulder patches, or depicts them with incorrect coloration or placement, introduces semantic mismatching that will degrade classifier performance. 

In light of this, this work explores the following key question:
\begin{qbox}
    \begin{center}
    \textit{
        \textbf{How to achieve both diversity and fine-grained fidelity in generative augmentation?}
    }
    \end{center}
\end{qbox}

To investigate the solution to the aforementioned question, we propose a \textit{\textbf{Hi}erarchically \textbf{G}uided \textbf{F}ine-grained \textbf{A}ugmentation method (\textbf{HiGFA})} that progressively incorporates conditions for controlling diversity and fidelity. 
Specifically, we jointly consider three conditions: 
\textbf{1)} \textit{Text guidance} for diversity, which defines the overall scene, style, and coarse object attributes; 
\textbf{2)} \textit{Diversity-enhanced Contour guidance} for structure fidelity, which ensures realistic object structure and integration with scene; 
\textbf{3)} \textit{Fine-grained classifier guidance} for category-specific fidelity, which ensures that the generated objects preserve the discriminative characteristics. Utilizing a conditional diffusion model~\cite{zhang2023addingcontrol} with three guidance, we can generate samples with both high diversity and strong category fidelity. 

To avoid conflicts among the three types of guidance, this paper further proposes a dynamic guidance mechanism. 
Relying on static weighting for these signals is inherently fragile. 
For example, strict adherence to contour guidance may enforce a subject's pose but conflict with background described in the text prompt, resulting in an unnatural ``cutout'' effect. 
Similarly, excessive classifier guidance can make image collapse.
Given that the optimal balance between these signals varies on an image-by-image basis, our approach leverages the observation that diffusion models generate images progressively, from coarse to fine\cite{li2024critical, raya2023spontaneous, benita2025designing, koulischer2025feedback}.

Consequently, we exploit this inherent temporal dynamic by controlling the guidance strengths with the classification confidences. 
The generation process begins with a warm-up phase where text and contour guidance predominantly shape low-frequency image components, establishing global structure.
Following this initial phase, classifier guidance is activated. 
The relative strengths of all three guidance signals, text, contour, and classifier, are then dynamically modulated based on the evolving classification confidence. 
Specifically, when high class fidelity is rapidly attained for a sample, indicated by high classifier confidence, the weight of the classifier guidance is progressively reduced, allowing text and contour cues to exert greater influence and thereby promote diversity. 
Conversely, for samples where achieving fine-grained fidelity proves more challenging, evidenced by lower confidence, the classifier guidance maintains a pronounced role, ensuring that category-specific details are accurately rendered until satisfactory confidence levels are reached.

In a nutshell, the contributions of this work are as follows:
\begin{itemize}
\item We propose \texttt{HiGFA}, a unified framework that synergistically integrates hierarchical guidance sources (text, contour, classifier) within the diffusion process.
\item We apply geometric transformations on contour maps to enhance sample diversity and implement a dynamic strategy to maintain a balance between global structure and fine-grained details.
\item We demonstrate that \texttt{HiGFA}  generates diverse, high-fidelity augmented images, leading to superior performance on FGVC tasks compared to existing generative augmentation methods.
\end{itemize}

\section{Related Work}

\subsubsection{Data Augmentation via Generative Models.} 
Generative models can synthesize novel and diverse examples that potentially lie outside the empirical support of the original training set, enriching the underlying data manifold.
Early work employed generative adversarial networks (GANs)~\cite{goodfellow2014generative} to synthesize additional samples for minority classes in imbalanced classification and fine-grained classification tasks~\cite{mariani2018bagan, zhang2021datasetgan}.
Several studies~\cite{hemmat2023feedback, trabucco2023effective, he2022synthetic, shama2024diffaug} have demonstrated that using diffusion models for data augmentation or expansion can improve performance on downstream classification tasks. 
Some of these works, such as~\cite{zhang2023expanding, he2022synthetic, dunlap2023diversify}, primarily rely on CLIP image features and CLIP-encoded text as conditions for data generation. 
However, due to the limited fine-grained perceptual capabilities of CLIP~\cite{hua2025openworldauc, hua2024reconboost, jiang2023hierarchical}, the generated samples often fail to accurately reflect fine-grained visual characteristics.
\citet{yangdistribution} proposed guiding the generation process using hierarchical prototypes extracted by feature extractors. 
However, the selection of prototypes in this method depends on hyperparameter choices, and although this approach enforces feature consistency, it does not explicitly encourage diversity in the generated samples. 
\citet{michaeli2024advancing} introduced the use of ControlNet to condition generation based on image structure and subject. 
However, for fine-grained image classification, preserving only the structure and subject is insufficient; the key performance factors typically lie in subtle visual details, rather than in maintaining a particular pose or object composition.
Our proposed method, \texttt{HiGFA}, synergistically integrates multiple guidance sources with a hierarchical and adaptive scheduling strategy.
We enhance diversity by applying random rotations and thin-plate spline interpolation to manipulate the edge maps input to ControlNet. 
To ensure consistency in the generated samples, we incorporate guidance from a fine-grained classifier.

\section{Methodology}
\label{sec:method}

\begin{figure*}[!t]
  \centering
  \includegraphics[width=1.99\columnwidth]{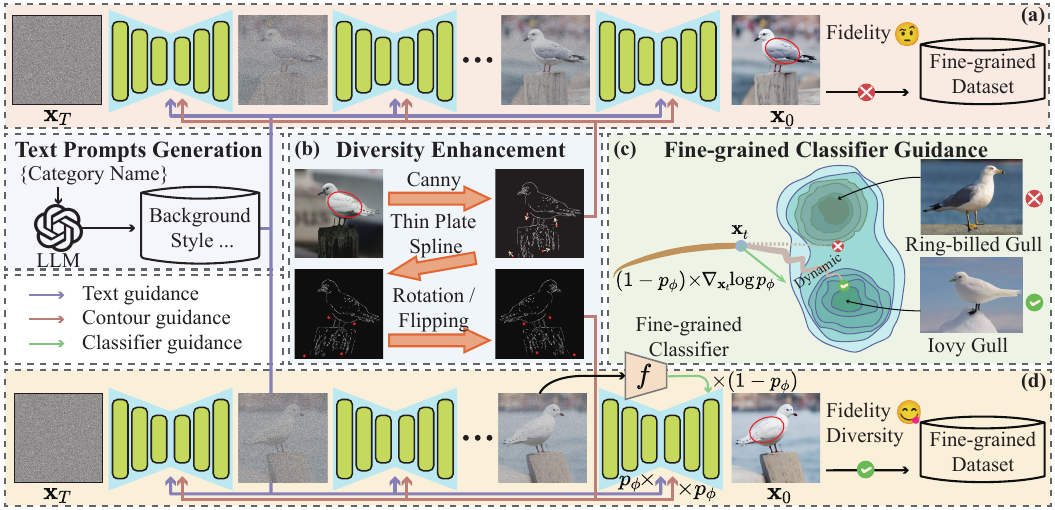} 
    \caption{
      \textbf{ An overview of our \texttt{HiGFA}:} 
      \textbf{a)} Existing diffusion-based augmentation method is insufficient to ensure category fidelity.
      \textbf{b)} Incorporating contour guidance helps preserve scene layout and object structure, but struggles with fine-grained category consistency and limit diversity. To enhance diversity, we apply diversity enhancement such as flipping, rotation, and thin-plate spline interpolation.
      \textbf{c)} To improve fine-grained category consistency, in the later stages, our adaptive strategy incorporates classifier guidance to balance the scale of all three guidance on a sample-wise level. Without classifier guidance, the generation process tends to converge toward the mean of the class distribution.
      \textbf{d)} Under the combination of hierarchical guidance and dynamic adjustment, our method generates augmented images with both high category fidelity and enhanced diversity.
      } 
  \label{fig:pipeline}
\end{figure*}

The primary objective of employing generative models for fine-grained data augmentation is to synthesize diverse images that faithfully represent the subtle characteristics of specific categories, thereby enhancing the performance of fine-grained classifiers.
In this section, we first introduce essential preliminaries on diffusion models and guided generation.
We then highlight the limitations of existing diffusion-based augmentation methods, particularly their inability to maintain category fidelity.
To address this issue, we propose a Hierarchically Guided Fine-grained Augmentation (\texttt{HiGFA}) method designed to preserve fidelity while simultaneously improving data diversity.

\subsection{Preliminaries}
\label{sec:preliminary}

\subsubsection{Diffusion Models.}
Diffusion models~\cite{ho2020denoising, song2021scorebased, lipman2023flow} are deep generative models that generate data by reversing a stochastic process with gradually adding noise.
Let $\mathbf{x}_0 \sim q(\mathbf{x}_0)$ be a sample from the true data distribution. 
The forward process is a Markov chain defined over $T$ timesteps, injecting Gaussian noise according to a variance schedule $\beta_t \in (0, 1)$:
$ q(\mathbf{x}_t | \mathbf{x}_{t-1}) = \mathcal{N}(\mathbf{x}_t; \sqrt{1 - \beta_t} \mathbf{x}_{t-1}, \beta_t \mathbf{I})$.
As $t \rightarrow T$, $\mathbf{x}_T$ approaches an isotropic Gaussian distribution $\mathcal{N}(\mathbf{0}, \mathbf{I})$.
The reverse process aims to denoise $\mathbf{x}_t$ back to $\mathbf{x}_{t-1}$, typically modeled by a neural network $\epsilon_\theta(\mathbf{x}_t, t)$ that predicts the noise added at step $t$.
The reverse transition $p_\theta(\mathbf{x}_{t-1} | \mathbf{x}_t)$ can be approximated as:
\begin{equation}
    p_\theta(\mathbf{x}_{t-1} \mid \mathbf{x}_t) = \mathcal{N}(
        \mathbf{x}_{t-1}; \boldsymbol{\mu}_{\boldsymbol{\theta}}(\mathbf{x}_t, t), \boldsymbol{\Sigma}_{\boldsymbol{\theta}}(\mathbf{x}_t, t)
    ),
\end{equation}
where $\boldsymbol{\mu}_{\boldsymbol{\theta}}$ and $\boldsymbol{\Sigma}_{\boldsymbol{\theta}}$ depend on the noise prediction $\epsilon_\theta(\mathbf{x}_t, t)$.
At inference, sampling starts from $\mathbf{x}_T \sim \mathcal{N}(\mathbf{0}, \mathbf{I})$ and iteratively applies the learned reverse process to generate a sample $\mathbf{x}_0$. The basic diffusion model described here is unconditional.

\subsubsection{Guided Diffusion.} 
Conditional generation can be achieved using guidance mechanisms. 
Classifier guidance~\cite{dhariwal2021diffusionbeatgans} uses a pretrained classifier $p_{\boldsymbol{\phi}}(y \mid \mathbf{x})$ to steer the sampling towards a target class $y$. 
The mean of the reverse step is adjusted using the classifier's gradient:
\begin{equation}
    \hat{\boldsymbol{\mu}}_{\boldsymbol{\theta}}(\mathbf{x}_{t}|y) = \boldsymbol{\mu}_{\boldsymbol{\theta}}(\mathbf{x}_{t}) + s \cdot \boldsymbol{\Sigma}_{\boldsymbol{\theta}}(\mathbf{x}_{t}) \nabla_{\mathbf{x}_{t}} \log p_{\boldsymbol{\phi}}(y|\mathbf{x}_{t}),
\end{equation}
where $s$ is the guidance scale.
Classifier-free guidance (CFG)~\cite{ho2022classifier} avoids external classifiers by jointly training a conditional model $\epsilon_{\theta}(\mathbf{x}_{t}, t, y)$ and an unconditional model $\epsilon_{\theta}(\mathbf{x}_{t}, t, \emptyset)$ (often by randomly dropping the conditioning $y$ during training). 
During sampling, the noise prediction is extrapolated from the conditional and unconditional predictions:
\begin{equation}\label{eq:classifier_free_guidance}
    \hat{\epsilon}_{\theta}(\mathbf{x}_{t} | y) = \epsilon_{\theta}(\mathbf{x}_{t} | \emptyset) + s \cdot (\epsilon_{\theta}(\mathbf{x}_{t} | y) - \epsilon_{\theta}(\mathbf{x}_{t} | \emptyset)),
\end{equation}
where $s$ is the guidance scale, $y$ is the condition (e.g., text prompt), and $\emptyset$ represents the null condition (e.g., empty string). 
CFG is highly effective and widely used in large-scale text-to-image models~\cite{rombach2022high, esser2024scaling, ramesh2022hierarchicaldalle2, podell2024sdxl}.

\subsection{Fidelity Challenge in Fine-grained Data Augmentation}
\label{sec:low_fidelity}

A primary challenge when using generative models for fine-grained data augmentation is ensuring the synthetic images accurately reflect the subtle, defining features of the target category. 
Standard text-to-image models, often guided by text prompts interpreted by models like CLIP~\cite{radford2021learningclip} or LLMs~\cite{raffel2020exploring}, struggle to capture these fine-grained distinctions, potentially leading to generated images with incorrect category attributes.

For instance, consider augmenting images for the ``{Ivory Gull}'', a seabird species characterized by entirely white plumage, distinguishing it from most other gulls which typically have gray mantles~\cite{Jobling2010TheHD}. 
Existing diffusion-based augmentation methods~\cite{dunlap2023diversify, he2022synthetic} typically rely on text guidance, often using classifier-free guidance (CFG). 
Even with the introduction of simple contour guidance~\cite{michaeli2024advancing}, these methods can only ensure contour similarity but fail to achieve the specificity required to preserve fine-grained key features.
Furthermore, maintaining fine-grained category definitions is an image-level challenge, and using fixed guidance weights makes it difficult to balance fidelity and diversity across all images.
This can result in generated samples that deviate from the true category characteristics, diminishing the effectiveness of the augmentation.

\subsection{Hierarchically Guided Fine-grained Augmentation}
\label{sec:ours_method}

The core limitation identified above is the lack of precise control at image-level over the generation process to enforce adherence to fine-grained category definitions. 
To address this, we propose Hierarchically Guided Fine-grained Augmentation ( \texttt{HiGFA} ). 
Our method is motivated by the observation that diffusion models tend to generate coarse, low-frequency information (e.g., overall structure, scene layout) during the early stages of the reverse sampling process and progressively refine high-frequency details in later stages~\cite{li2024critical, benita2025designing}. 
\texttt{HiGFA} leverages this temporal dynamic by applying multiple forms of guidance with varying strengths across the sampling process, transitioning from coarse-grained control to fine-grained feature enforcement:

\begin{itemize}
    \item \textbf{Text Guidance:} Primarily drives \textit{diversity} by leveraging text prompts via classifier-free guidance to define varied global compositions, scenes, and styles. Its strong initial influence establishes a semantically coherent foundation for image generation.
    \item \textbf{Diversity-enhanced Contour Guidance:} Enforces structural \textit{fidelity} by applying ControlNet~\cite{zhang2023addingcontrol} with transformed edge maps. This ensures realistic structures and seamless integration into scene, while contour transformations further support \textit{diversity} in object pose and layout.
    \item \textbf{Fine-grained Classifier Guidance:} Improves fine-grained \textit{fidelity} by refining category-specific visual details during the final sampling stages. This tailored classifier guidance accurately captures subtle, discriminative features critical to the target class.
\end{itemize}

\subsubsection{Text Guidance.}
Given a text prompt $u$, specifying a desired scene, general object category, and optionally an artistic style, we use standard classifier-free guidance (CFG, Eq. \ref{eq:classifier_free_guidance}) to provide coarse-grained control. 
Following~\cite{michaeli2024advancing} prompts are generated using category names and scene descriptions, with optional style modifiers  (e.g., ``photorealistic'', ``oil painting'') introduced randomly for diversity. 
The strength of CFG is controlled by a scale parameter $s_{\mathrm{cfg}}(t)$.
This scale remains constant during initial sampling steps ($t = T$ down to $N_s$) to robustly establish the global structure and style, and is dynamically adjusted during later steps.

\subsubsection{Diversity-enhanced Contour Guidance.}
To maintain structural fidelity with real images from the dataset, we incorporate contour guidance using ControlNet~\cite{zhang2023addingcontrol}. 
For a reference image $\mathbf{x}$ from the dataset, we extract its Canny edge map $\mathbf{x}_c$ and condition the diffusion process on it via ControlNet. 
To enhance diversity beyond the reference image's specific pose and structure, we apply transformations to the Canny edge map $\mathbf{x}_c$ before feeding.
For datasets with rigid objects (e.g., cars, airplanes), we apply random horizontal flipping and random rotation (e.g., within $\pm 15^\circ$).
For datasets with non-rigid objects (e.g., birds, dogs), we additionally apply thin-plate spline (TPS) warping~\cite{bookstein1989principal} to simulate plausible deformations.
Specifically, we divide the edge map into patches, randomly select five edge-containing patches, and perturb a point in each to create a target map for TPS.
These augmentations encourage pose diversity.
ControlNet's influence is governed by a conditioning strength parameter $s_{\mathrm{ctl}}(t)$, which remains constant during the initial phase and is dynamically modulated for $t < N_s$.

\begin{table*}[t]
\centering

\begin{tabular}{rlcccccc}
\toprule
\multicolumn{1}{c}{\multirow{2}[4]{*}{\textbf{Type}}} & \multicolumn{1}{l}{\multirow{2}[4]{*}{\textbf{Methods}}} & \multicolumn{3}{c}{{\textbf{Rigid Objects}}} & \multicolumn{3}{c}{{\textbf{Non-Rigid Objects}}} \\
\cmidrule{3-8}          &       & \textbf{Aircraft} & \textbf{Cars}  & \textbf{CompCars} & \textbf{CUB}   & \textbf{Dogs}  & \textbf{DTD} \\
\midrule
\multicolumn{1}{r}{\multirow{6}[2]{*}{Traditional}} 
& NoAug~\cite{rao2021counterfactual}            & \orangee{82.6}  & \orangef{91.8}  & \orangef{67.0}  & \bluee{81.5}  & \bluee{84.1}  & \bluef{68.5}  \\
& FRAug~\cite{rao2021counterfactual}             & \orangec{83.2}  & \orangee{92.1}  & \orangee{69.4}  & \bluec{82.0}  & \blued{84.2}  & \bluec{69.2}  \\
& CALAug~\cite{rao2021counterfactual}           & \orangeb{84.9}  & \oranged{92.4}  & \orangee{70.5}  & \blueb{\underline{82.5}}  & \bluec{84.6}  & \blueb{\underline{69.7}}  \\
& RandAug~\cite{cubuk2020randaugment}           & \orangec{83.7}  & \oranged{92.6}  & \orangec{72.5}  & \bluee{81.5}  & \blued{84.2}  & \bluec{69.3}  \\
& AutoAug~\cite{autoaug}           & \orangec{84.7}  & \orangeb{92.9}  & \orangeb{73.2}  & \bluec{82.1}  & \bluec{84.6}  & \bluec{68.7}  \\
& RandEarse~\cite{zhong2020random}        & \orangee{82.5}  & \orangee{92.3}  & \oranged{71.4}  & \bluef{81.2}  & \bluef{83.8}  & \bluee{69.5}  \\
& CutMix~\cite{yun2019cutmix}            & \oranged{83.0}  & \orangef{91.7}  & \orangef{67.3}  & \bluee{81.8}  & \bluec{84.5}  & \bluec{69.2}  \\
\midrule
\multicolumn{1}{r}{\multirow{6}[2]{*}{Generative}} 
& Real Guidance~\cite{he2022synthetic}     & \oranged{83.2}  & \orangee{92.0}  & \orangec{72.8}  & \blueb{82.2}  & \bluee{83.1}  & \bluef{68.2}  \\
& ALIA~\cite{dunlap2023diversify}              & \orangee{83.0}  & \orangef{91.9}  & \orangec{72.2}  & \bluef{80.8}  & \bluec{83.8}  & \bluee{68.9}  \\
& DiffuseMix~\cite{islam2024diffusemix}        & \orangef{81.8}  & \orangee{92.3}  & \oranged{72.0}  & \bluee{81.6}  & \bluec{84.0}  & \bluee{68.7}  \\
& DistDiff~\cite{yangdistribution}          & \orangef{82.4}  & \orangee{92.1}  & \oranged{71.8}  & \bluef{80.9}  & \blued{84.2}  & \bluee{68.8}  \\
& SaSPA~\cite{michaeli2024advancing}             & \orangeb{\underline{85.2}}  & \orangeb{\underline{93.0}}  & \orangeb{\underline{74.4}}  & \bluec{81.7}  & \blueb{\underline{84.9}}  & \blueb{\underline{69.7}}  \\
& \textbf{Ours}     & \orangea{\textbf{86.1}} & \orangea{\textbf{93.6}} & \orangea{\textbf{75.2}} & \bluea{\textbf{82.7}} & \bluea{\textbf{85.8}} & \bluea{\textbf{70.8}} \\
\bottomrule
\end{tabular}%
\caption{
    Performance comparison of data augmentation methods using the CAL (ResNet50) classifier across rigid and non-rigid datasets. 
    Results are reported as top-1 Accuracy (\%). 
    Best results are in \textbf{bold}, runner-up \underline{underlined}, respectively.
    }
\label{tab:fgvc_results}
\end{table*}

\begin{table*}[!ht]
\centering

\begin{tabular}{cllllllll}
\toprule
\multicolumn{1}{l}{\textbf{Dataset}} & \textbf{Methods} & \textbf{No Aug} & \textbf{FRAug} & \textbf{CAL-Aug} & \textbf{RandAug} & \textbf{AutoAug} & \textbf{RandEarse} & \textbf{CutMix} \\
\midrule
\multirow{2}[2]{*}{Aircraft} & Baseline & 82.6  & 83.2  & 84.9  & 83.7  & 84.7  & 82.5  & 83.0  \\
        & Ours      
        & 85.8\textbf{ (+3.2)}
        & 86.1\textbf{ (+2.9)}
        & 87.5\textbf{ (+2.6)}
        & 86.6\textbf{ (+2.9)}
        & 86.5\textbf{ (+1.8)}
        & 85.4\textbf{ (+2.9)}
        & 85.8\textbf{ (+2.8)}
        \\
\midrule
\multirow{2}[2]{*}{Dogs} & Baseline & 84.1  & 84.2  & 84.6  & 84.2  & 84.6  & 83.8  & 84.5 \\
        & Ours  
        & 85.1\textbf{ (+1.0)} 
        & 85.8\textbf{ (+1.6)} 
        & 86.2\textbf{ (+1.6)} 
        & 85.7\textbf{ (+1.5)} 
        & 85.3\textbf{ (+0.7)} 
        & 84.8\textbf{ (+1.0)} 
        & 85.2\textbf{ (+0.7)} \\
\bottomrule
\end{tabular}%
\caption{
    Comparison of traditional augmentation methods on FGVC Aircraft and Stanford Dogs. Our method demonstrates strong compatibility with conventional data augmentation techniques.
}
\label{tab:fgvc_results_waug}
\end{table*}

\subsubsection{Fine-grained Classifier Guidance.}
To enforce category-specific details, we employ a fine-grained classifier $p_{\phi}(y \mid \mathbf{x}')$, trained on the original dataset, as a domain-specific expert to ensure the generated images accurately exhibit category-specific details.
This guidance is \textbf{activated only} during the later sampling stages (for $t < N_s$), when structural elements are largely resolved and noise levels are reduced.
Let $p_{\phi}(y | \mathbf{x}')$ be the fine-grained classifier, where $\mathbf{x}'$ is an image and $y$ is the target fine-grained label. 
Since the classifier operates on images rather than latent variables, we estimate the corresponding clean image $\hat{\mathbf{x}}_0$ from the noisy latent $\mathbf{x}_t$ using the standard denoising prediction:
\begin{equation}\label{eq:predict_x0}
    \hat{\mathbf{x}}_0(\mathbf{x}_t, t) = \frac{\mathbf{x}_t - \sqrt{1 - \bar{\alpha}_t} \epsilon_\theta(\mathbf{x}_t, t)}{\sqrt{\bar{\alpha}_t}},
\end{equation}
where $\bar{\alpha}_t = \prod_{i=1}^t (1 - \beta_i)$.
The predicted latent $\hat{\mathbf{x}}_0$ is then decoded to an image $\mathbf{x}'_t$ using the VAE decoder $\mathfrak{D}$:
\begin{equation}\label{eq:decode_x0}
    \mathbf{x}'_{t} = \mathfrak{D}(\hat{\mathbf{x}}_0(\mathbf{x}_t, t)).
\end{equation}

The fine-grained guidance adjusts the noise prediction $\epsilon_\theta$, which already incorporates CFG and ControlNet, denoted $\epsilon_\theta$. 
For $t < N_s$, the adjusted noise prediction is given by:
\begin{equation}\label{eq:classifier_guidance_ours}
    \tilde{\epsilon}_{\theta}(\mathbf{x}_{t}, t) = \epsilon_{\theta}(\mathbf{x}_{t}, t) - s_{\mathrm{cls}}(t) \cdot \sigma_t \cdot \nabla_{\mathbf{x}_{t}} \log p_{\phi}(y \mid \mathbf{x}'_{t}),
\end{equation}
where $s_{\mathrm{cls}}(t)$ is the classifier guidance scale, and $\sigma_t = 1.0$ is the gradient scaling factor, following standard practice~\cite{nichol2021glide}.

Crucially, accounting for the varying difficulty in extracting fine-grained features from images within the same dataset is vital.
For example, images with clean backgrounds require minimal classifier guidance to avoid image collapse due to excessive guidance, whereas images with cluttered backgrounds or multiple objects demand stronger guidance to preserve class fidelity.
Therefore, we propose a dynamic guidance adjustment mechanism that adaptively modulates the strength of guidance based on classifier predictions, aligning it with the complexity of individual images.
All guidance strengths are dynamically orchestrated in two phases, delineated by a threshold timestep $N_s$.
Let $p_{\phi}(y \mid \mathbf{x}'_t)$ denote the classifier's confidence in the target class at time $t$.
During the early steps ($t \geq N_s$), the objective is to establish global structure and style. 
Both $s_{\mathrm{cfg}}$ and $s_{\mathrm{ctl}}$ are fixed, and fine-grained guidance remains inactive.
In the later steps ($t < N_s$), the objective shifts to refining fine-grained details. 
The guidance strengths are updated as follows:
\begin{equation}
    \begin{aligned}
    s_{\mathrm{cfg}}(t) &= s_{\mathrm{cfg}}(T) \cdot p_{\phi}(y \mid \mathbf{x}'_{t-1}), \\
    s_{\mathrm{ctl}}(t) &= s_{\mathrm{ctl}}(T) \cdot p_{\phi}(y \mid \mathbf{x}'_{t-1}), \\
    s_{\mathrm{cls}}(t) &= s_{\mathrm{cls}}(N_s) \cdot \left(1 - p_{\phi}(y \mid \mathbf{x}'_t)\right),
    \end{aligned}
\end{equation}
where $s_{\mathrm{cfg}}(T)$ and $s_{\mathrm{ctl}}(T)$ are the initial CFG and ControlNet scales, respectively, and $s_{\mathrm{cls}}(N_s)$ is the initial classifier guidance scale.
When the classifier exhibits uncertain (i.e., $p_{\phi}(y \mid \mathbf{x}'_t)$ is low), the guidance is amplified to steer generation more strongly toward the target class.
Conversely, when classifier confidence is high, the guidance is attenuated to prevent overcorrection, allowing the base generation process to continue with minimal interference. 
This adaptive mechanism enhances stability and mitigates overfitting to classifier biases.


\begin{figure*}[!t]
  \centering
  \includegraphics[width=1.0\linewidth]{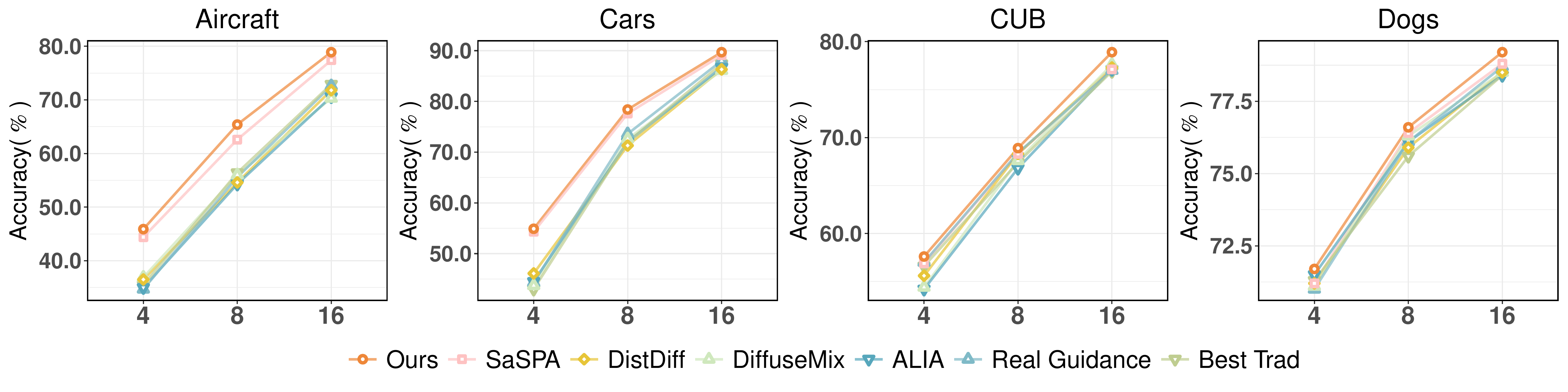} 
  \caption{
    Performance of methods on the FGVC Aircraft, Stanford Cars, CUB, and Stanford Dogs under few-shot settings. The results indicate that our method remains effective even when the classifier is not very reliable.
  }
  \label{fig:few_shot_results}
\end{figure*}

\section{Experiments}
\label{sec:experiments}

\subsection{Experiment Setup}
\label{subsec:Setting}
\subsubsection{Dataset.} 
Six FGVC benchmarks are used to evaluate including FGVC Aircraft~\cite{maji2013fine}, CUB200-2011~\cite{reed2016learning}, Stanford Cars~\cite{krause20133d}, Stanford Dogs~\cite{khosla2011novel}, CompCars~\cite{yang2015large} and DTD~\cite{cimpoi2014describing}. 
The FGVC Aircraft, Stanford Cars, and CompCars datasets primarily evaluate the performance of our method on rigid objects, whereas the CUB200-2011, Stanford Dogs, and DTD datasets assess its effectiveness on non-rigid objects.

\subsubsection{Comparison Methods.}
We compare our method against the following competitors:
\begin{itemize}
\item \textbf{Traditional Augmentation Methods:} 
FRAug(random horizontal flipping and random rotation),
CALAug~\cite{rao2021counterfactual}(random flipping, random cropping, and color jittering),
RandAug~\cite{cubuk2020randaugment}, 
AutoAug~\cite{autoaug}, 
RandEarse~\cite{zhong2020random}, 
CutMix~\cite{yun2019cutmix}.
\item \textbf{Diffusion Based Augmentation Methods:} 
Real Guidance~\cite{he2022synthetic}, a method that uses a low translation strength to preserve fidelity to the original image;
ALIA~\cite{dunlap2023diversify}, a method that uses both the original dataset caption and a GPT-generated summary as textual guidance;
DiffuseMix~\cite{islam2024diffusemix}, an image-to-image method that leverages diffusion models to address category ambiguity introduced by mixup;
DistDiff~\cite{yangdistribution}, a method that uses hierarchical prototypes to guide sample generation, ensuring alignment with the distribution of the original dataset;
SaSPA~\cite{michaeli2024advancing}, a method designed for FGVC, utilizing ControlNet to preserve the structure and subject of the original images.
\end{itemize}

\subsection{Fine-grained Visual Classification}
\label{subsec:fgvc_results}
\subsubsection{Implementation Details.}
For all generative model-based methods, we generate two augmented images per original image. 
Across all datasets, the training set consists of $40\%$ generated images and $60\%$ original images, except for the CUB dataset, where the augmentation ratio is reduced to $20\%$. 
We add the random horizontal flipping and random rotation to all diffusion-based methods.

\subsubsection{Results.}
\begin{figure*}[!t]
  \centering
  \includegraphics[width=0.98\linewidth]{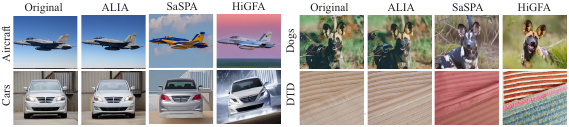} 
  \caption{
  Qualitative comparisons between images generated by \texttt{HiGFA} and some baseline generative augmentation methods for fine-grained categories. 
    These comparisons highlight \texttt{HiGFA}'s ability to generate diverse images while also better preserving subtle category-defining details.
    The complete results are provided in Appendix.
  }
  \label{fig:qua_vis}
\end{figure*}

The results presented in \Cref{tab:fgvc_results} demonstrate the effectiveness of our proposed augmentation method across a diverse range of fine-grained visual benchmarks, encompassing both rigid and non-rigid object categories.
\textbf{1)} First, our method consistently achieves state-of-the-art performance, securing the top accuracy score on all six datasets. 
This highlights the robustness and general applicability of our approach. 
\textbf{2)} Compared to using no augmentation, our method provides substantial gains, ranging from $+1.2\%$ on CUB to $+8.2\%$ on CompCars, indicating the significant benefit of the generated data.
In addition, most data augmentation methods based on generative models are superior to traditional data augmentation methods.
These results suggest that the generative approach introduces more beneficial variations than standard geometric and photometric transformations as shown in \Cref{fig:qua_vis}.
\textbf{3)} Notably, compared to SaSPA, which is specifically designed for FGVC and often the strongest generative competitor, our method shows considerable improvements.
\textbf{4)} \Cref{tab:fgvc_results_waug} presents the results of integrating our method with commonly used traditional data augmentation techniques. 
The performance improvements observed demonstrate that our approach significantly benefits from these combinations, further highlighting its robustness.
This consistent advantage over other state-of-the-art generative methods underscores the effectiveness of our specific augmentation strategy.

\subsection{Few-shot Learning}
\label{sec:few_shot}
\subsubsection{Implementation Details.}
To evaluate the performance of our method under limited data conditions, we conduct experiments in the few-shot learning setting. 
Specifically, we select four datasets, Aircraft, Cars, CUB, and Dogs, along with several representative comparison methods.
And we consider 4-shot, 8-shot, and 16-shot scenarios. 
In this setting, the guidance classifier is replaced with a ResNet50 model \textbf{trained solely} on the available $k$-shot samples.
Additionally, we increased the augmentation ratio to 0.6.

\subsubsection{Results.}
As shown in \Cref{fig:few_shot_results}, our method consistently outperforms other baselines in few data setting. 
This demonstrates that even when the guidance classifier is unreliable, our dynamic adjustment mechanism effectively mitigates over-perturbation, ensuring that the generated images are not overly influenced by the classifier.

\subsection{Ablation Study}
\label{sec:mai_ablation}

\subsubsection{Hierarchical Guidance.} 
We select FGVC Aircraft, Stanford Cars, CUB, and DTD, to perform ablation experiments on hierarchical guidance.
\Cref{tab:ablation_hierarchy} shows that each component contributes incrementally to performance improvement.
Notably, on non-rigid datasets, omitting classifier guidance results in performance degradation may below the baseline. 
This indicates that, for non-rigid categories, text and contour guidance alone are insufficient for the generative model to produce discriminative features.
Overall, incorporating fine-grained classifier guidance yields substantial performance gains, highlighting its critical role in generating high-fidelity images for fine-grained visual classification tasks.

\begin{table}[!t]
\centering
\setlength{\tabcolsep}{1.2mm}{
\begin{tabular}{ccccccc}
\toprule
Text & Contour & Classifier & Aircraft & Cars & CUB & DTD \\
\midrule
\checkmark &            &            & 83.8 & 91.9  & 78.9 & 63.5 \\
\checkmark & \checkmark &            & 85.3 & 92.5  & 80.2 & 66.6 \\
\checkmark & \checkmark & \checkmark & 86.1 & 93.6  & 82.7 & 70.8 \\
\bottomrule
\end{tabular}%
}
\caption{
    Ablation study on hierarchical guidance.
}
\label{tab:ablation_hierarchy}

\end{table}

\begin{figure*}[t]
    \centering
    \begin{subfigure}{0.27\linewidth}
        \includegraphics[width=\linewidth]{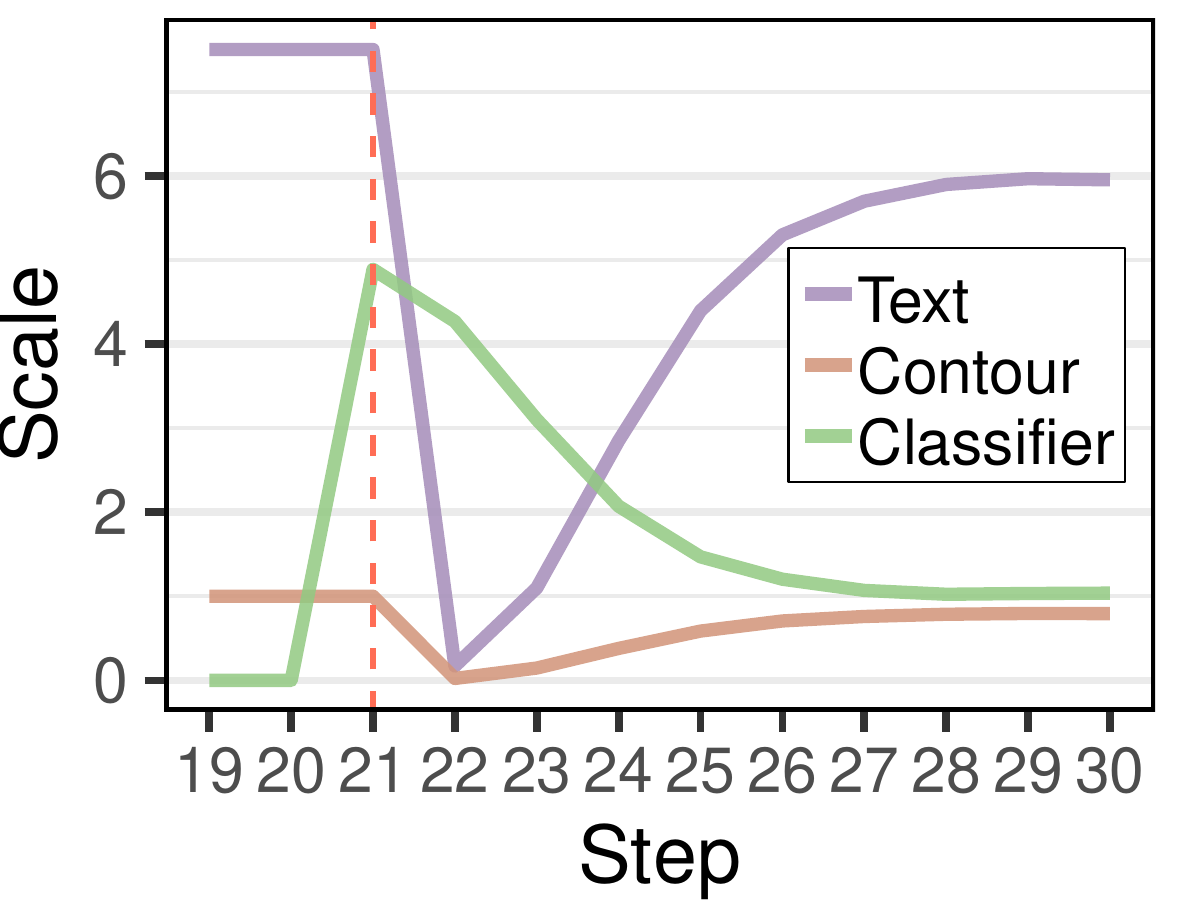}
        \caption{\footnotesize{Average evolution.}}
        \label{fig:Feature_guidance_adaptive}
    \end{subfigure}
    \begin{subfigure}{0.27\linewidth}
        \includegraphics[width=\linewidth]{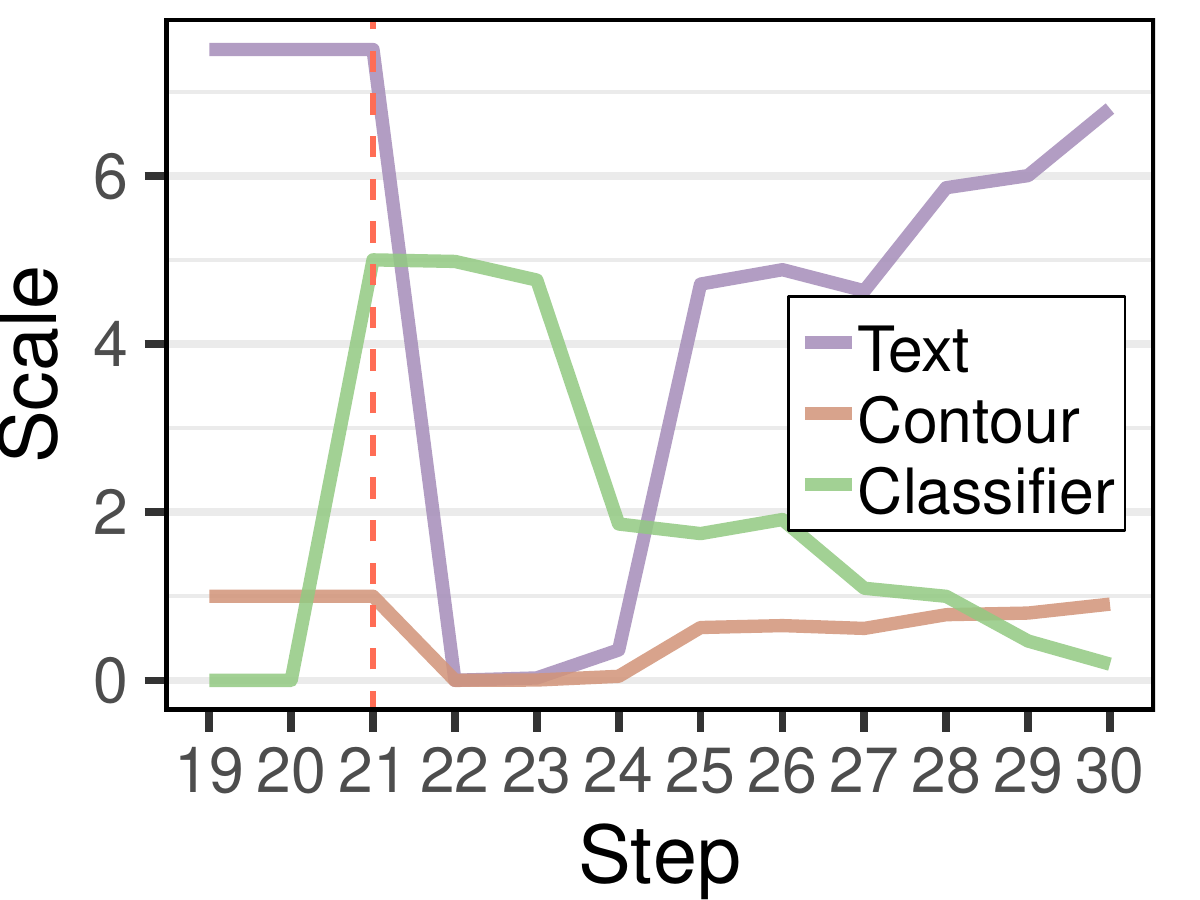}
        \caption{\footnotesize{Easy sample.}}
        \label{fig:Feature_guidance_adaptive_simple}
    \end{subfigure}
    \begin{subfigure}{0.27\linewidth}
        \includegraphics[width=\linewidth]{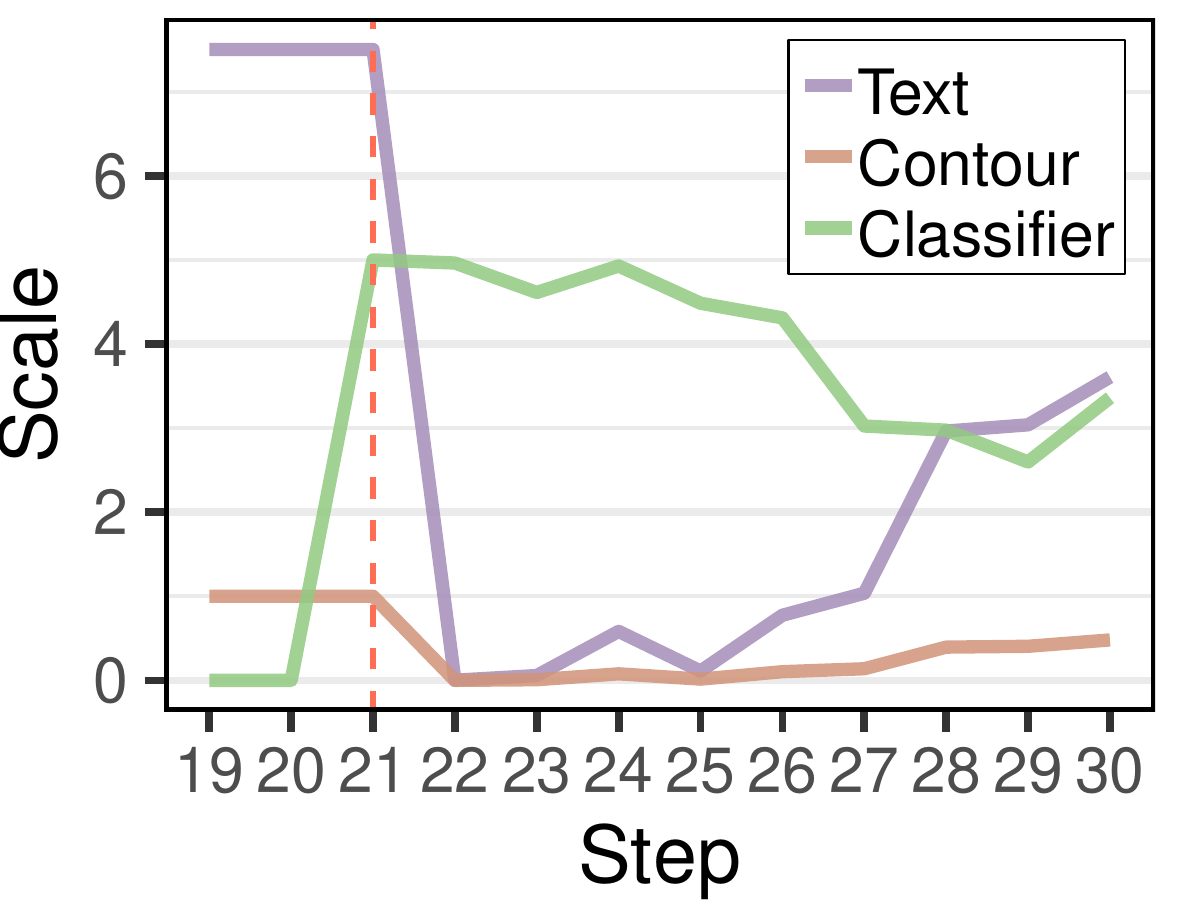}
        \caption{\footnotesize{Hard sample.}}
        \label{fig:Feature_guidance_adaptive_hard}
    \end{subfigure}
    \caption{
    Evolution of the guidance scale during generation. 
    The red dashed line indicates the activation of classifier guidance and the dynamic mechanism.
    Easy samples with clean backgrounds need only brief classifier signals to guide generation, while hard samples with complex backgrounds and conflicting prompts require extended guidance for generation, highlighting the sample-wise adaptability of our dynamic guidance.
    }
    \label{fig: Ablation_study}
\end{figure*}

\subsubsection{Classifier Guidance Scale and Step.} 

\begin{figure}[!t]
    \centering
    \subfloat[Classifier guidance scale.]{
        \includegraphics[width=0.49\linewidth]{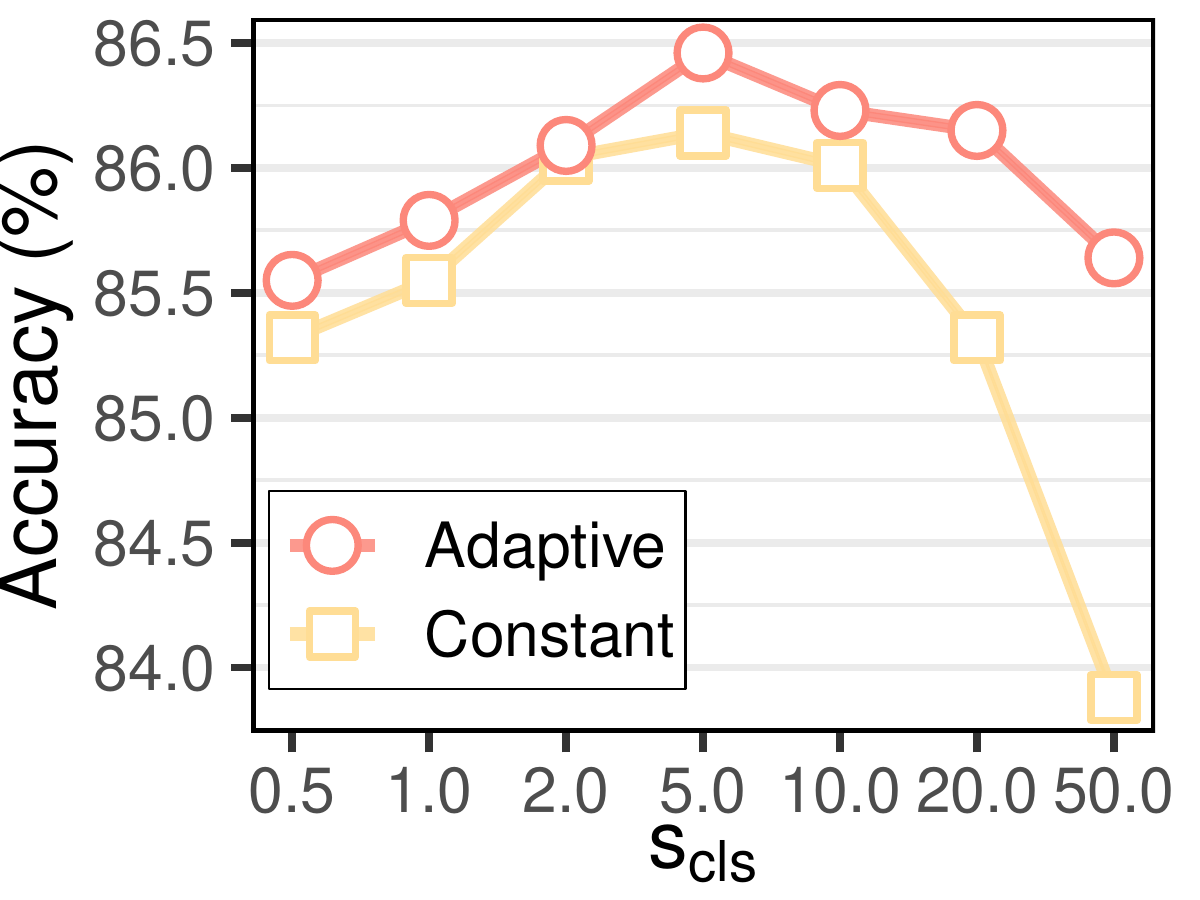}
        \label{fig:Feature_guidance_scale}
    }
    \subfloat[Classifier guidance start step.]{
        \includegraphics[width=0.49\linewidth]{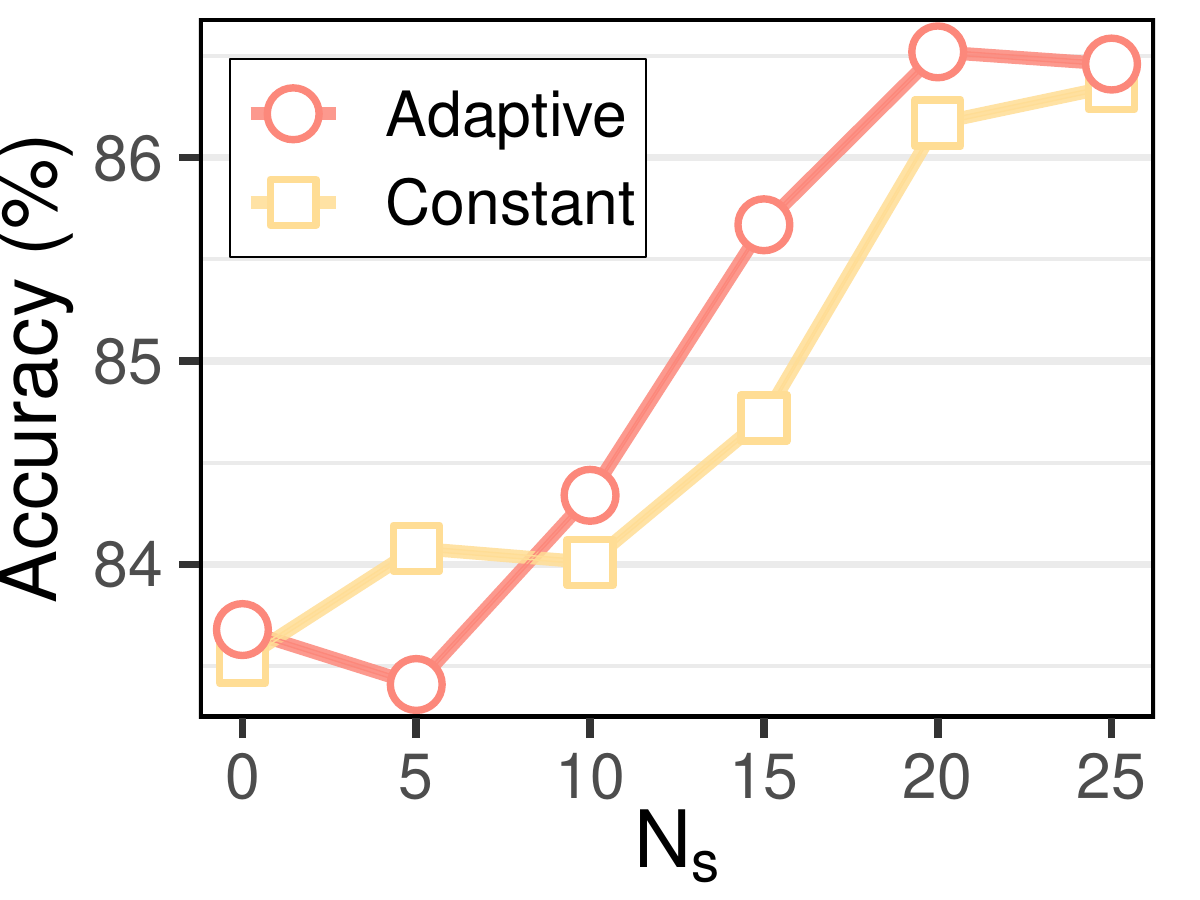}
        \label{fig:Feature_guidance_start_step}
    }
    \caption{
    Ablation study on guidance scale, step and adaptive strategy.
    Enabling classifier guidance in the later stages of diffusion improves efficiency and enhances the robustness.
    }
    \label{fig:Ablation_study_on_scale_and_step}
\end{figure}

We conduct ablation studies on the FGVC Aircraft dataset to investigate the effects of the classifier guidance scale and the classifier guidance start step.
\Cref{fig:Feature_guidance_scale} examines the impact of the classifier guidance scale $s_{\mathrm{cls}}$, with the start step fixed at $N_s = 21$.

When $s_{\mathrm{cls}}$ is small, the influence of classifier guidance is limited, causing the diffusion model to rely more heavily on textual and contour cues.
Conversely, when $s_{\mathrm{cls}}$ is too large, classifier guidance becomes overly dominant, which may result in partial image collapse and semantic distortion, ultimately degrading classifier performance.
\Cref{fig:Feature_guidance_start_step} analyzes the effect of the classifier guidance start step $N_s$, with the scale fixed at $s_{\mathrm{cls}} = 5$.
Results indicate that applying classifier guidance during the later stages of the diffusion process yields superior outcomes.
Introducing guidance too early, when background structures and object contours are still forming, can be harmful, as classifier predictions at this stage are often meaningless.
Early intervention may disrupt both textual and contour guidance. 

\subsubsection{Adaptive Strategy.} 
\Cref{fig:Ablation_study_on_scale_and_step} present an ablation study evaluating the dynamic strategy.
Without this strategy, increasing the guidance scale significantly degrades classification performance.
Incorporating it reduces sensitivity to the hyperparameter $s_{\mathrm{cls}}$ and improves the performance and stability of \texttt{HiGFA}.
\Cref{fig:Feature_guidance_adaptive} shows the average evolution of the three guidance during inference on the FGVC Aircraft.
At the onset of classifier guidance, both text and contour guidance sharply decline. 
As inference process continues, all three guidance gradually stabilize.
\Cref{fig:Feature_guidance_adaptive_simple} and \Cref{fig:Feature_guidance_adaptive_hard} illustrate the evolution of the guidance scale for an easy and a hard sample, respectively. 
For the easy sample, images with clean backgrounds and the main subject minimally affected by the surroundings, a brief but strong classifier signal suffices to guide generation toward the target class, after which text and contour guidance dominate.
In contrast, the hard sample, which feature complex backgrounds and conflicts between text prompt and the main subject, requires prolonged classifier guidance to progressively align the output with the target distribution.

\section{Conclusion}
\label{sec:discussion}

This paper introduces Hierarchically Guided Fine-grained Augmentation (\texttt{HiGFA}) for generating images for fine-grained data augmentation with diffusion models. 
Existing guidance methods often fail to capture subtle, category-defining features, limiting their augmentation effectiveness. 
\texttt{HiGFA} solves this problem by combining multiple guidance sources: text, diversity-enhanced contours, and a classifier. 
It adjusts the strength of each source over time to match the diffusion model's coarse-to-fine image generation process. 
Early stages focus on global structure and style using text and contour guidance, while later stages employ an adaptive, confidence-modulated fine-grained classifier guidance to precisely enforce category-specific features on a cleaner image estimate. 
Experiments on several FGVC datasets demonstrate the effectiveness of \texttt{HiGFA}.

\section{Acknowledgments}
This work was supported in part by National Natural Science Foundation of China: 62525212, 62236008, 62441232, U21B2038, and U23B2051, in part by Youth Innovation Promotion Association CAS, in part by the Strategic Priority Research Program of the Chinese Academy of Sciences, Grant No. XDB0680201, in part by the China National Postdoctoral Program for Innovative Talents, Grant No. BX20250377.

\bibliography{aaai2026}

@String(PAMI  = {IEEE TPAMI})

@String(CVPR  = {CVPR})

@String(ICCV  = {ICCV})

@String(NeurIPS = {NeurIPS})

@String(ICML  = {ICML})

@String(ICLR  = {ICLR})

@String(AAAI  = {AAAI})

@String(PAMI  = {IEEE Transactions on Pattern Analysis and Machine Intelligence})

@String(CVPR  = {Proceedings of the IEEE/CVF Conference on Computer Vision and Pattern Recognition})

@String(ICCV  = {International Conference on Computer Vision})

@String(NeurIPS = {Annual Conference on Neural Information Processing Systems})

@String(ICML  = {International Conference on Machine Learning})

@String(ICLR  = {International Conference on Learning Representations})

@String(AAAI  = {Proceedings of the AAAI Conference on Artificial Intelligence})

@article{maji2013fine,
  title={Fine-grained visual classification of aircraft},
  author={Maji, Subhransu and Rahtu, Esa and Kannala, Juho and Blaschko, Matthew and Vedaldi, Andrea},
  journal={arXiv preprint arXiv:1306.5151},
  year={2013}
}

@inproceedings{krause20133d,
  title={3d object representations for fine-grained categorization},
  author={Krause, Jonathan and Stark, Michael and Deng, Jia and Fei-Fei, Li},
  booktitle={Proceedings of the IEEE International Conference on Computer Vision Workshops},
  pages={554--561},
  year={2013}
}

@inproceedings{reed2016learning,
  title={Learning deep representations of fine-grained visual descriptions},
  author={Reed, Scott and Akata, Zeynep and Lee, Honglak and Schiele, Bernt},
  booktitle=CVPR,
  pages={49--58},
  year={2016}
}

@inproceedings{yang2015large,
  title={A large-scale car dataset for fine-grained categorization and verification},
  author={Yang, Linjie and Luo, Ping and Change Loy, Chen and Tang, Xiaoou},
  booktitle=CVPR,
  pages={3973--3981},
  year={2015}
}

@inproceedings{cimpoi2014describing,
  title={Describing textures in the wild},
  author={Cimpoi, Mircea and Maji, Subhransu and Kokkinos, Iasonas and Mohamed, Sammy and Vedaldi, Andrea},
  booktitle=CVPR,
  pages={3606--3613},
  year={2014}
}

@inproceedings{khosla2011novel,
  title={Novel dataset for fine-grained image categorization: Stanford dogs},
  author={Khosla, Aditya and Jayadevaprakash, Nityananda and Yao, Bangpeng and Li, Fei-Fei},
  booktitle={Proceedings of the IEEE Conference on Computer Vision and Pattern Recognition Workshops},
  volume={2},
  year={2011}
}

@article{achiam2023gpt,
  title={Gpt-4 technical report},
  author={Achiam, Josh and Adler, Steven and Agarwal, Sandhini and Ahmad, Lama and Akkaya, Ilge and Aleman, Florencia Leoni and Almeida, Diogo and Altenschmidt, Janko and Altman, Sam and Anadkat, Shyamal and others},
  journal={arXiv preprint arXiv:2303.08774},
  year={2023}
}

@inproceedings{rombach2022high,
  title={High-resolution image synthesis with latent diffusion models},
  author={Rombach, Robin and Blattmann, Andreas and Lorenz, Dominik and Esser, Patrick and Ommer, Bj{\"o}rn},
  booktitle=CVPR,
  pages={10684--10695},
  year={2022}
}

@article{nichol2021glide,
  title={Glide: Towards photorealistic image generation and editing with text-guided diffusion models},
  author={Nichol, Alex and Dhariwal, Prafulla and Ramesh, Aditya and Shyam, Pranav and Mishkin, Pamela and McGrew, Bob and Sutskever, Ilya and Chen, Mark},
  journal={arXiv preprint arXiv:2112.10741},
  year={2021}
}

@inproceedings{dhariwal2021diffusionbeatgans,
  title={Diffusion models beat gans on image synthesis},
  author={Dhariwal, Prafulla and Nichol, Alexander},
  booktitle=NeurIPS,
  volume={34},
  pages={8780--8794},
  year={2021}
}

@article{ho2022classifier,
  title={Classifier-Free Diffusion Guidance},
  author={Ho, Jonathan and Salimans, Tim},
  journal={arXiv preprint arXiv:2207.12598},
  year={2022}
}

@article{ramesh2022hierarchicaldalle2,
  title={Hierarchical text-conditional image generation with clip latents},
  author={Ramesh, Aditya and Dhariwal, Prafulla and Nichol, Alex and Chu, Casey and Chen, Mark},
  journal={arXiv preprint arXiv:2204.06125},
  volume={1},
  number={2},
  pages={3},
  year={2022}
}

@inproceedings{saharia2022photorealisticimagen,
  title={Photorealistic text-to-image diffusion models with deep language understanding},
  author={Saharia, Chitwan and Chan, William and Saxena, Saurabh and Li, Lala and Whang, Jay and Denton, Emily L and Ghasemipour, Kamyar and Gontijo Lopes, Raphael and Karagol Ayan, Burcu and Salimans, Tim and others},
  booktitle=NeurIPS,
  volume={35},
  pages={36479--36494},
  year={2022}
}

@inproceedings{zhang2023addingcontrol,
  title={Adding conditional control to text-to-image diffusion models},
  author={Zhang, Lvmin and Rao, Anyi and Agrawala, Maneesh},
  booktitle=ICCV,
  pages={3836--3847},
  year={2023}
}

@inproceedings{he2022synthetic,
  title={Is Synthetic Data from Generative Models Ready for Image Recognition?},
  author={Ruifei He and Shuyang Sun and Xin Yu and Chuhui Xue and Wenqing Zhang and Philip Torr and Song Bai and Xiaojuan Qi},
  booktitle=ICLR,
  year={2023},
}

@inproceedings{trabucco2023effective,
  title={Effective Data Augmentation with Diffusion Models},
  author={Brandon Trabucco and Kyle Doherty and Max Gurinas and Ruslan Salakhutdinov},
  booktitle=ICLR,
  year={2024},
}

@inproceedings{dunlap2023diversify,
  title={Diversify Your Vision Datasets with Automatic Diffusion-Based Augmentation},
  author={Lisa Dunlap and Alyssa Umino and Han Zhang and Jiezhi Yang and Joseph Gonzalez and Trevor Darrell},
  booktitle=NeurIPS,
  year={2023},
}

@inproceedings{li2024blipdiffusion,
  title={Blip-diffusion: Pre-trained subject representation for controllable text-to-image generation and editing},
  author={Li, Dongxu and Li, Junnan and Hoi, Steven},
  booktitle=NeurIPS,
  volume={36},
  year={2024}
}

@inproceedings{yun2019cutmix,
  title={Cutmix: Regularization strategy to train strong classifiers with localizable features},
  author={Yun, Sangdoo and Han, Dongyoon and Oh, Seong Joon and Chun, Sanghyuk and Choe, Junsuk and Yoo, Youngjoon},
  booktitle=ICCV,
  pages={6023--6032},
  year={2019}
}

@inproceedings{radford2021learningclip,
  title={Learning transferable visual models from natural language supervision},
  author={Radford, Alec and Kim, Jong Wook and Hallacy, Chris and Ramesh, Aditya and Goh, Gabriel and Agarwal, Sandhini and Sastry, Girish and Askell, Amanda and Mishkin, Pamela and Clark, Jack and others},
  booktitle=ICML,
  pages={8748--8763},
  year={2021},
}

@misc{paszke2019pytorch,
  title={PyTorch: An Imperative Style, High-Performance Deep Learning Library},
  author={Adam Paszke and Sam Gross and Francisco Massa and Adam Lerer and James Bradbury and Gregory Chanan and Trevor Killeen and Zeming Lin and Natalia Gimelshein and Luca Antiga and Alban Desmaison and Andreas K{\"o}pf and Edward Yang and Zachary DeVito and Martin Raison and Alykhan Tejani and Sasank Chilamkurthy and Benoit Steiner and Lu Fang and Junjie Bai and Soumith Chintala},
  booktitle={Advances in Neural Information Processing Systems 32},
  editor={H. Wallach and H. Larochelle and A. Beygelzimer and F. d\textquotesingle Alch\'{e}-Buc and E. Fox and R. Garnett},
  pages={8024--8035},
  year={2019},
  publisher={Curran Associates, Inc.}
}

@inproceedings{rao2021counterfactual,
  title={Counterfactual attention learning for fine-grained visual categorization and re-identification},
  author={Rao, Yongming and Chen, Guangyi and Lu, Jiwen and Zhou, Jie},
  booktitle=ICCV,
  pages={1025--1034},
  year={2021}
}

@misc{diffusers,
  author = {Patrick von Platen and Suraj Patil and Anton Lozhkov and Pedro Cuenca and Nathan Lambert and Yehao Li and Mishig Davaakhuu and Aedan S. Culotta and Camilo Rodrigues},
  title = {Diffusers: State-of-the-art diffusion models},
  year = {2022},
  publisher = {GitHub},
  howpublished = {\url{https://github.com/huggingface/diffusers}},
  note = {Accessed: 2023-05-10}
}

@inproceedings{cubuk2020randaugment,
  title={Randaugment: Practical automated data augmentation with a reduced search space},
  author={Cubuk, Ekin D and Zoph, Barret and Shlens, Jonathon and Le, Quoc V},
  booktitle={Proceedings of the IEEE/CVF Conference on Computer Vision and Pattern Recognition Workshops},
  pages={702--703},
  year={2020}
}

@inproceedings{he2016deepresnet,
  title={Deep residual learning for image recognition},
  author={He, Kaiming and Zhang, Xiangyu and Ren, Shaoqing and Sun, Jian},
  booktitle=CVPR,
  pages={770--778},
  year={2016}
}

@inproceedings{song2020denoising,
    title={Denoising Diffusion Implicit Models},
    author={Jiaming Song and Chenlin Meng and Stefano Ermon},
    booktitle=ICLR,
    year={2021},
}

@inproceedings{zhang2021datasetgan,
  title={Datasetgan: Efficient labeled data factory with minimal human effort},
  author={Zhang, Yuxuan and Ling, Huan and Gao, Jun and Yin, Kangxue and Lafleche, Jean-Francois and Barriuso, Adela and Torralba, Antonio and Fidler, Sanja},
  booktitle=CVPR,
  pages={10145--10155},
  year={2021}
}

@inproceedings{ho2020denoising,
  title={Denoising diffusion probabilistic models},
  author={Ho, Jonathan and Jain, Ajay and Abbeel, Pieter},
  booktitle=NeurIPS,
  volume={33},
  pages={6840--6851},
  year={2020}
}

@inproceedings{michaeli2024advancing,
    title={Advancing Fine-Grained Classification by Structure and Subject Preserving Augmentation},
    author={Eyal Michaeli and Ohad Fried},
    booktitle=NeurIPS,
    year={2024},
}

@inproceedings{song2021scorebased,
title={Score-Based Generative Modeling through Stochastic Differential Equations},
author={Yang Song and Jascha Sohl-Dickstein and Diederik P Kingma and Abhishek Kumar and Stefano Ermon and Ben Poole},
booktitle=ICLR,
year={2021},
}

@inproceedings{lipman2023flow,
title={Flow Matching for Generative Modeling},
author={Yaron Lipman and Ricky T. Q. Chen and Heli Ben-Hamu and Maximilian Nickel and Matthew Le},
booktitle=ICLR,
year={2023},
}

@inproceedings{esser2024scaling,
  title={Scaling rectified flow transformers for high-resolution image synthesis},
  author={Esser, Patrick and Kulal, Sumith and Blattmann, Andreas and Entezari, Rahim and M{\"u}ller, Jonas and Saini, Harry and Levi, Yam and Lorenz, Dominik and Sauer, Axel and Boesel, Frederic and others},
  booktitle=ICML,
  year={2024}
}

@inproceedings{podell2024sdxl,
  title={{SDXL}: Improving Latent Diffusion Models for High-Resolution Image Synthesis},
  author={Dustin Podell and Zion English and Kyle Lacey and Andreas Blattmann and Tim Dockhorn and Jonas M{\"u}ller and Joe Penna and Robin Rombach},
  booktitle={The Twelfth International Conference on Learning Representations},
  year={2024}
}

@book{Jobling2010TheHD,
  title={The Helm Dictionary of Scientific Bird Names: From Aalge to Zusii},
  author={James A. Jobling},
  year={2010},
  publisher={London: Christopher Helm},
}

@article{raffel2020exploring,
  title={Exploring the limits of transfer learning with a unified text-to-text transformer},
  author={Raffel, Colin and Shazeer, Noam and Roberts, Adam and Lee, Katherine and Narang, Sharan and Matena, Michael and Zhou, Yanqi and Li, Wei and Liu, Peter J},
  journal={Journal of machine learning research},
  volume={21},
  number={140},
  pages={1--67},
  year={2020}
}

@inproceedings{li2024critical,
  title={Critical windows: non-asymptotic theory for feature emergence in diffusion models},
  author={Li, Marvin and Chen, Sitan},
  booktitle={International Conference on Machine Learning},
  pages={27474--27498},
  year={2024},
  organization={PMLR}
}

@inproceedings{raya2023spontaneous,
  title={Spontaneous symmetry breaking in generative diffusion models},
  author={Raya, Gabriel and Ambrogioni, Luca},
  booktitle={Advances in Neural Information Processing Systems},
  volume={36},
  pages={66377--66389},
  year={2023}
}

@article{benita2025designing,
  title={Designing Scheduling for Diffusion Models via Spectral Analysis},
  author={Benita, Roi and Elad, Michael and Keshet, Joseph},
  journal={arXiv preprint arXiv:2502.00180},
  year={2025}
}

@article{bookstein1989principal,
  title={Principal warps: Thin-plate splines and the decomposition of deformations},
  author={Bookstein, Fred L.},
  journal=PAMI,
  volume={11},
  number={6},
  pages={567--585},
  year={1989},
  publisher={IEEE}
}

@inproceedings{mariani2018bagan,
  title={BAGAN: Data Augmentation with Balancing GAN},
  author={Mariani, Giovanni and Scheidegger, Florian and Istrate, Roxana and Bekas, Costas and Malossi, Cristiano},
  booktitle={International Conference on Machine Learning},
  year={2018}
}

@inproceedings{goodfellow2014generative,
  title={Generative adversarial nets},
  author={Goodfellow, Ian J and Pouget-Abadie, Jean and Mirza, Mehdi and Xu, Bing and Warde-Farley, David and Ozair, Sherjil and Courville, Aaron and Bengio, Yoshua},
  booktitle=NeurIPS,
  volume={27},
  year={2014}
}

@inproceedings{zhang2023expanding,
  title={Expanding small-scale datasets with guided imagination},
  author={Zhang, Yifan and Zhou, Daquan and Hooi, Bryan and Wang, Kai and Feng, Jiashi},
  booktitle=NeurIPS,
  volume={36},
  pages={76558--76618},
  year={2023}
}

@inproceedings{shama2024diffaug,
  title={DiffAug: A Diffuse-and-Denoise Augmentation for Training Robust Classifiers},
  author={Shama Sastry, Chandramouli and Dumpala, Sri Harsha and Oore, Sageev},
  booktitle=NeurIPS,
  volume={37},
  pages={20745--20785},
  year={2024}
}

@inproceedings{yangdistribution,
  title={Distribution-Aware Data Expansion with Diffusion Models},
  author={Yang, Ling and Yong, Jun-Hai and Yin, Hongzhi and Jiang, Jiawei and Xiao, Meng and Zhang, Wentao and Wang, Bin and others},
  booktitle=NeurIPS,
  year={2024}
}

@inproceedings{islam2024diffusemix,
  title={Diffusemix: Label-preserving data augmentation with diffusion models},
  author={Islam, Khawar and Zaheer, Muhammad Zaigham and Mahmood, Arif and Nandakumar, Karthik},
  booktitle={Proceedings of the IEEE/CVF Conference on Computer Vision and Pattern Recognition},
  pages={27621--27630},
  year={2024}
}

@inproceedings{zhong2020random,
  title={Random erasing data augmentation},
  author={Zhong, Zhun and Zheng, Liang and Kang, Guoliang and Li, Shaozi and Yang, Yi},
  booktitle=AAAI,
  volume={34},
  pages={13001--13008},
  year={2020}
}

@inproceedings{autoaug,
  title	= {AutoAugment: Learning Augmentation Policies from Data},author	= {Ekin Dogus Cubuk and Barret Zoph and Dandelion Mane and Vijay Vasudevan and Quoc V. Le},
  year	= {2019},
  booktitle=CVPR, 
}

@inproceedings{he2022transfg,
  title={Transfg: A transformer architecture for fine-grained recognition},
  author={He, Ju and Chen, Jie-Neng and Liu, Shuai and Kortylewski, Adam and Yang, Cheng and Bai, Yutong and Wang, Changhu},
  booktitle=AAAI,
  volume={36},
  pages={852--860},
  year={2022}
}

@article{hemmat2023feedback,
  title={Feedback-guided data synthesis for imbalanced classification},
  author={Hemmat, Reyhane Askari and Pezeshki, Mohammad and Bordes, Florian and Drozdzal, Michal and Romero-Soriano, Adriana},
  journal={arXiv preprint arXiv:2310.00158},
  year={2023}
}

@article{koulischer2025feedback,
  title={Feedback guidance of diffusion models},
  author={Koulischer, Felix and Handke, Florian and Deleu, Johannes and Demeester, Thomas and Ambrogioni, Luca},
  journal={arXiv preprint arXiv:2506.06085},
  year={2025}
}

@inproceedings{hua2024reconboost,
    title={ReconBoost: Boosting Can Achieve Modality Reconcilement}, 
    author={Cong Hua and Qianqian Xu and Shilong Bao and Zhiyong Yang and Qingming Huang},
    booktitle={International Conference on Machine Learning},
    pages = {19573--19597},
    year={2024}
}

@inproceedings{hua2025openworldauc,
  title={OpenworldAUC: Towards Unified Evaluation and Optimization for Open-world Prompt Tuning},
  author={Hua, Cong and Xu, Qianqian and Yang, Zhiyong and Wang, Zitai and Bao, Shilong and Huang, Qingming},
  booktitle={International Conference on Machine Learning},
  year={2025}
}

@article{jiang2023hierarchical,
  title={Hierarchical set-to-set representation for 3-d cross-modal retrieval},
  author={Jiang, Yu and Hua, Cong and Feng, Yifan and Gao, Yue},
  journal={IEEE Transactions on Neural Networks and Learning Systems},
  year={2023},
  publisher={IEEE}
}

\newpage
\section{Appendix}

\subsection{Additional Related Work}
\label{app:related_work}

\subsubsection{Text-to-Image Diffusion Models.} 
The field of text-to-image synthesis has seen remarkable progress driven by diffusion models coupled with large-scale language encoders like CLIP~\cite{radford2021learningclip} and T5~\cite{raffel2020exploring}. 
Models such as GLIDE~\cite{nichol2021glide}, \DALLE 2~\cite{ramesh2022hierarchicaldalle2}, Imagen~\cite{saharia2022photorealisticimagen}, and Stable Diffusion~\cite{rombach2022high, podell2024sdxl} can generate highly realistic and diverse images from complex textual descriptions. 
These models typically employ classifier-free guidance (CFG)~\cite{ho2022classifier} to align the generated image with the input prompt. 
However, their ability to capture subtle, fine-grained visual details solely from text prompts is often limited. 
Text prompts are insufficient to precisely describe the distinguishing features of closely related fine-grained categories. 
\texttt{HiGFA} utilizes a standard text-to-image diffusion model as the backbone but recognizes the limitations of relying solely on text prompts for fine-grained fidelity, supplementing CFG with diversity-enhanced contour and classifier guidance.

\subsubsection{Conditional Diffusion Models.}
Controlling the output of diffusion models under multiple conditions remains an active area of research.
Classifier guidance~\cite{dhariwal2021diffusionbeatgans} is an early approach, using gradients from a pretrained classifier to steer sampling. 
Although conceptually straightforward, it requires a robust classifier trained on noisy images and is limited to generating specific categories.
Classifier-free guidance (CFG)~\cite{ho2022classifier} eliminates the need for an external classifier and became the predominant technique for text-to-image generation due to its flexibility and effectiveness.
More recently, methods such as ControlNet~\cite{zhang2023addingcontrol} introduce spatial conditioning, such as edge maps, segmentation maps, or human poses, through auxiliary modules that inject information into the diffusion model's backbone.
Our proposed method, \texttt{HiGFA}, integrates these multiple guidance: standard CFG for scene and style, ControlNet for structural contour guidance, and an adapted form of classifier guidance specifically for fine-grained features. 
The key novelty lies not only in combining three types of guidance, but in applying them hierarchically with dynamically adjusted strengths mechanism synchronized with the diffusion timesteps, reflecting the coarse-to-fine generation process.

\subsection{Experiments}
\label{app_section:experiments}

\subsubsection{Datasets Details.}
\label{app:dataset}
We used the official validation sets where available. 
For datasets lacking a predefined validation set (e.g., CUB200-2011, Stanford Dogs), we randomly held out approximately $33\%$ of the training data for validation. 
Data augmentation, both traditional and generative, was applied exclusively to the training images. 
To ensure consistency with comparison methods, we used only exterior car parts from the CompCars dataset, focusing specifically on categories such as headlights, taillights, fog lights, and air intakes.
Dataset statistics are summarized in \Cref{tab:dataset_details}.

\begin{table}[!h]   
    \centering
    \setlength{\tabcolsep}{1mm}{
        \begin{tabular}{cccccc}
            \toprule
            \textbf{Type} & \textbf{Datasets}        & \textbf{Classes} & \textbf{Train}   & \textbf{Val}    & \textbf{Test} \\
            \midrule
            \multirow{3}{*}{Rigid}
            & FGVC Aircraft   & 100    & 3,334 & 3,333 & 3,333 \\
            & Stanford Cars   & 196    & 5,457 & 2,687 & 8,041 \\
            & CompCars        & 431    & 3,733 & 1,838 & 4,683 \\
            \midrule
            \multirow{3}{*}{Non-Rigid}
            & CUB200-2011     & 200    & 4,016 & 1,978 & 5,794 \\
            & Stanford Dogs   & 120    & 8,040 & 3,960 & 8,580 \\
            & DTD             & 47     & 1,880 & 1,880 & 1,880 \\
            \bottomrule
        \end{tabular}%
    }
    \caption{Statistics of the datasets.}
    \label{tab:dataset_details}
\end{table}

\subsubsection{Comparison Methods Details.}
\label{app:comparsion}
Below are details of the comparison methods used in our experiments:
\begin{itemize}
    \item \textbf{Traditional Augmentation Methods:} 
    \begin{itemize}
        \item[*] \textbf{FRAug}: Applies random horizontal flipping and random rotation.
        \item[*] \textbf{CALAug}~\cite{rao2021counterfactual}: Applies random horizontal flipping, random cropping, and color jittering.
        \item[*] \textbf{RandAug}~\cite{cubuk2020randaugment}: Employs a fixed number of randomly selected augmentation operations, such as rotation, brightness and color jittering.
        \item[*] \textbf{AutoAug}~\cite{autoaug}: Learns augmentation policies from data, applying sequences of operations with specific probabilities and magnitudes.
        \item[*] \textbf{RandErase}~\cite{zhong2020random}: Randomly selects a rectangle area in an image and erases its pixels.
        \item[*] \textbf{CutMix}~\cite{yun2019cutmix}: Replaces a random region of an input image with a patch from another training image, mixing their labels proportionally.
    \end{itemize}
    \item \textbf{Diffusion Based Augmentation Methods:} 
    \begin{itemize}
        \item[*] \textbf{Real Guidance}~\cite{he2022synthetic}: An image-to-image method using diffusion with a low translation strength ($s = 0.15$) to generate variations while preserving fidelity to the original image.
        \item[*] \textbf{ALIA}~\cite{dunlap2023diversify}: An image-to-image method guided by both the original image caption and a GPT-generated summary description as textual prompts.
        \item[*] \textbf{DiffuseMix}~\cite{islam2024diffusemix}: An image-to-image method that uses diffusion models to refine images generated by mixup, aiming to reduce category ambiguity.
        \item[*] \textbf{DistDiff}~\cite{yangdistribution}: A text-to-image method guided by hierarchical prototypes extracted from a feature extractor, ensuring generated samples align with the original data distribution.
        \item[*] \textbf{SaSPA}~\cite{michaeli2024advancing}: A text-to-image method designed for FGVC that uses ControlNet conditioned on structural features to preserve the pose and structure of the subject from a reference image.
    \end{itemize}
\end{itemize}

\subsubsection{Compute Resources.}
\label{app:compute_resources}
All experiments were conducted on NVIDIA RTX 4090 GPUs. 
During the image generation process, GPU memory usage did not exceed 16 GB.
For classifier training, the CAL method with a ResNet-50 backbone required no more than 10 GB of GPU memory, whereas the TransFG method with a ViT-B backbone consumed up to 24 GB.
On average, generating a single image required approximately 5.5 seconds, and the total training time per dataset was under 2 hours.
For large-scale datasets, parallel processing can be utilized to further reduce runtime.
Standard server-grade CPUs and sufficient system RAM were used for data loading and auxiliary processing, but these were not computational bottlenecks.

\subsubsection{More Implementation Details}
\label{app:more_Implementation_details}
Our implementation uses PyTorch~\cite{paszke2019pytorch} and the Diffusers library~\cite{diffusers}.
For all diffusion-based methods, including ours, we use Stable Diffusion v1.5~\cite{rombach2022high} as the base generative model, except for the SaSPA, which is built upon BLIP-Diffusion~\cite{li2024blipdiffusion}.
We employed the DDIM sampler~\cite{song2020denoising} with 30 inference steps.
Classifier-Free Guidance (CFG)~\cite{ho2022classifier} is used with a scale of $s_{\mathrm{cfg}}=7.5$. 
For methods using ControlNet~\cite{zhang2023addingcontrol}, the conditioning scale is set to $s_{\mathrm{ctl}}=1.0$.
Generated images have their shorter dimension resized to 512 pixels. 
For each training image, we generate two augmented versions using two distinct, randomly selected prompts from a predefined set. 
We use the prompts provided by the SaSPA repository~\cite{michaeli2024advancing} for consistency. 
For Stanford Dogs, where prompts were unavailable, we generate them using GPT-4~\cite{achiam2023gpt}, following the SaSPA method.

ControlNet is conditioned on Canny edge maps extracted using thresholds of 120 (low) and 200 (high), respectively.
For Thin Plate Spline (TPS) warping~\cite{bookstein1989principal} applied to Canny maps for non-rigid objects, control points are selected by first detecting corners using OpenCV. 
From the corners within patches containing edges, we select $k=5$ points that are maximally distant from each other to define the warp. 
The warped edge map is binarized using a threshold of 100.

For classifier-guided methods \texttt{HiGFA} and DistDiff~\cite{yangdistribution}, the guidance classifier $p_{\phi}$ in the main experiments 
is CAL~\cite{rao2021counterfactual} with a ResNet-101~\cite{he2016deepresnet} backbone. 
For the few-shot experiments 
, this is replaced with a ResNet-50 backbone trained on the $k$-shot data. 
The guidance activation timestep was $N_s = 20$, and the total number of diffusion steps is $30$. 
The initial classifier guidance scale was $s_{\mathrm{cls}}(N_s) = 5$.

In the main fine-grained classification experiments 
, the downstream evaluation classifier was CAL~\cite{rao2021counterfactual} with a ResNet-50~\cite{he2016deepresnet} backbone, initialized with ImageNet-pretrained weights. 
Input images were resized to $224 \times 224$. 
We used the SGD optimizer with a learning rate schedule and trained for 140 epochs. 
For few-shot learning 
, we trained for 100 epochs.
For the experiments using a ViT-based classifier 
, we employed TransFG~\cite{he2022transfg} (ViT-B/14) with an input resolution of $448 \times 448$ pixels. 
We used the SGD optimizer (Adam for CompCars) and trained for 10,000 steps. 
All classifiers were initialized with ImageNet-pretrained weights. 
All reported metrics are averages over three runs with different random seeds.

\subsubsection{Effect of Augmentation Ratio on Performance.}
\begin{figure*}[h]
    \centering
    \begin{subfigure}{0.3\linewidth}
        \includegraphics[width=\linewidth]{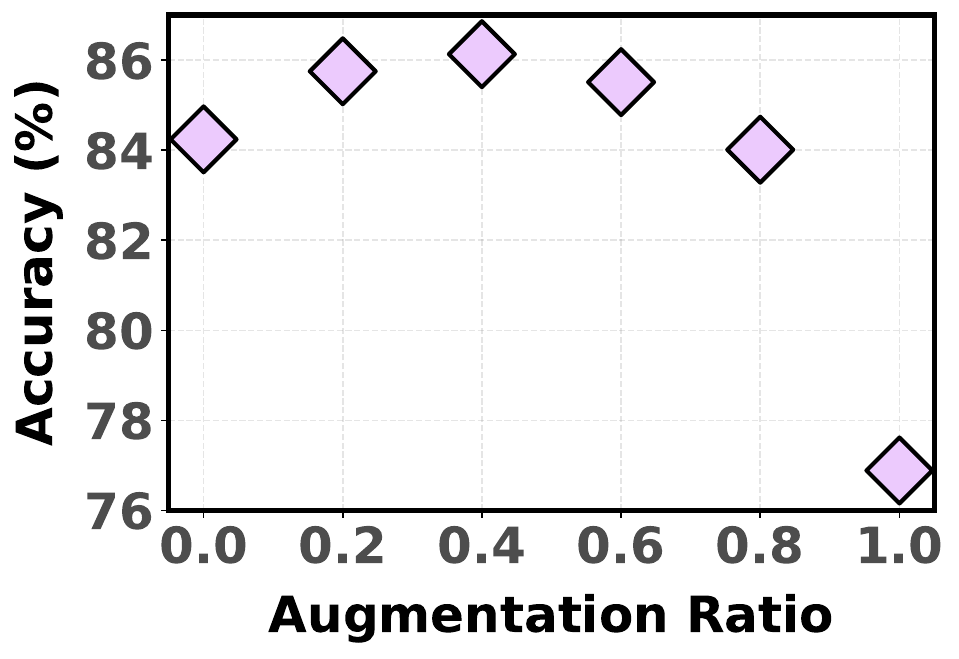}
        \caption{FGVC Aircraft}
        \label{fig:planes_ratio}
    \end{subfigure}
    \begin{subfigure}{0.3\linewidth}
        \includegraphics[width=\linewidth]{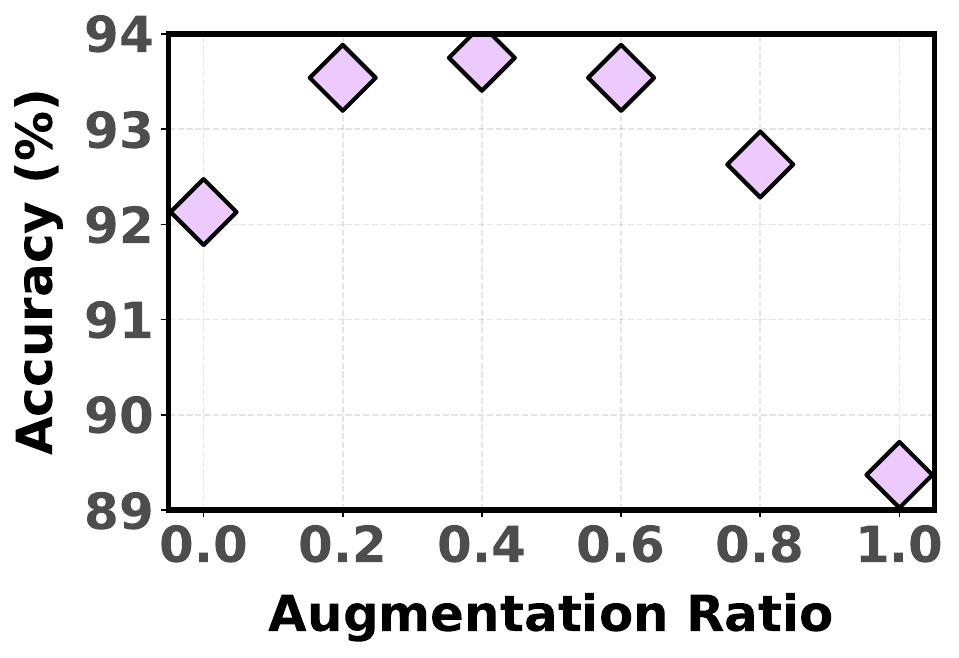}
        \caption{Stanford Cars}
        \label{fig:cars_ratio}
    \end{subfigure}
    \begin{subfigure}{0.3\linewidth}
        \includegraphics[width=\linewidth]{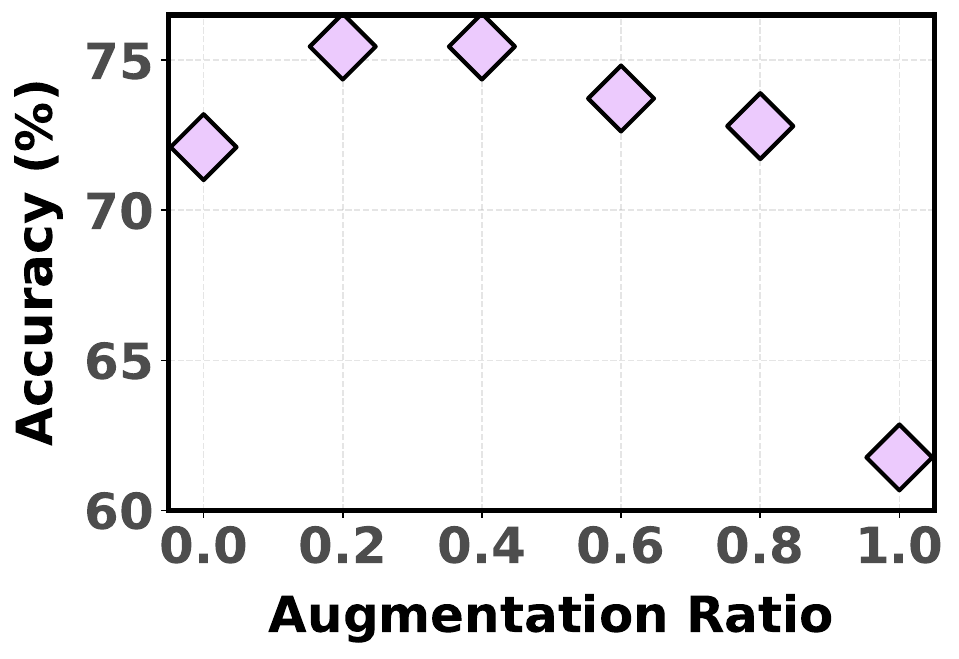}
        \caption{CompCars}
        \label{fig:compcars_ratio}
    \end{subfigure}
    \\
    \begin{subfigure}{0.3\linewidth}
            \includegraphics[width=\linewidth]{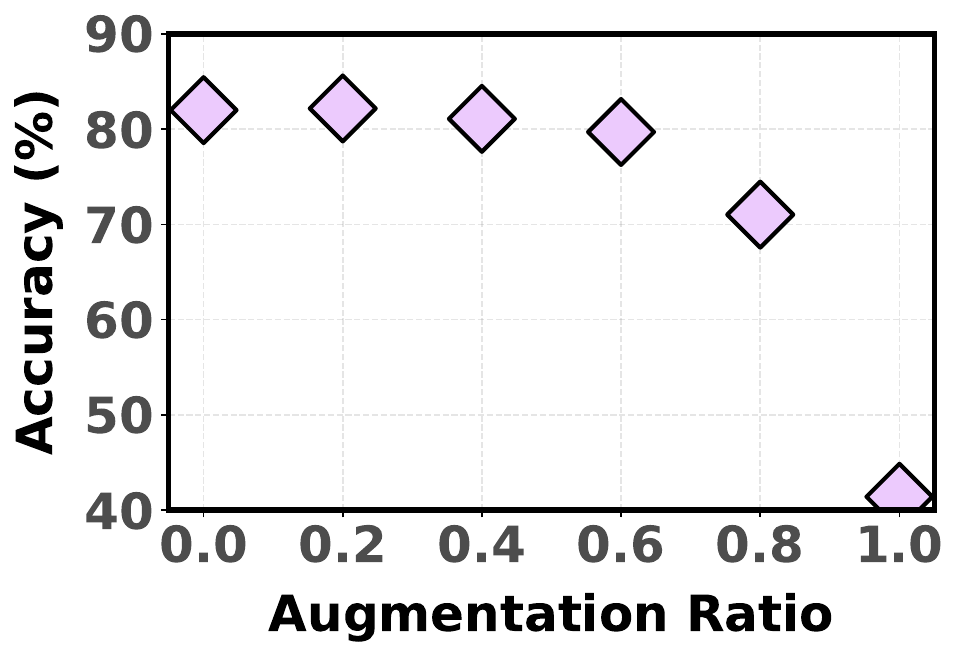}
        \caption{CUB200-2011}
        \label{fig:cub_ratio}
    \end{subfigure}
    \begin{subfigure}{0.3\linewidth}
        \includegraphics[width=\linewidth]{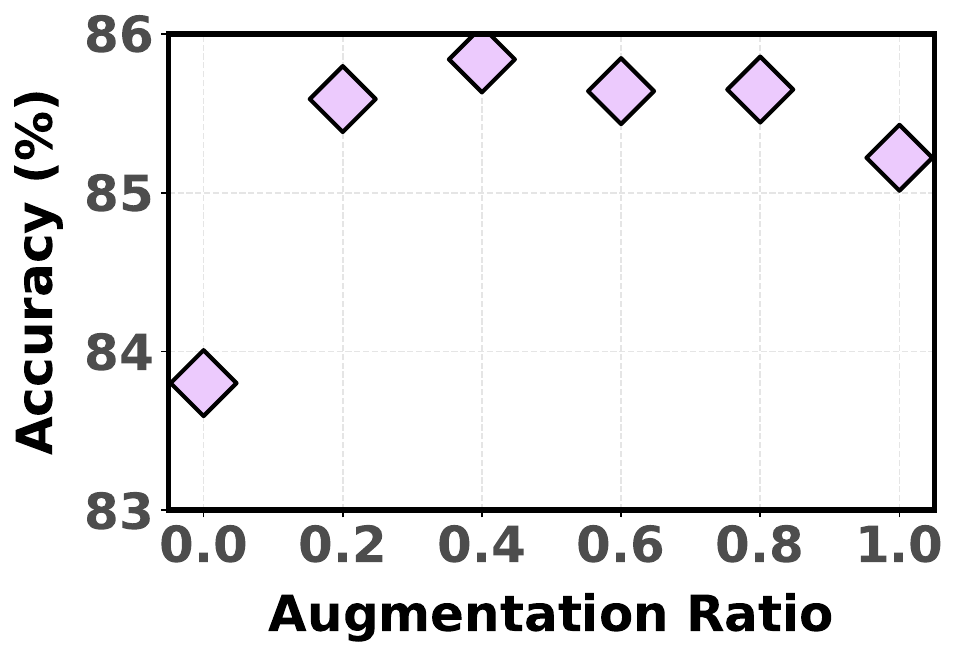}
        \caption{Stanford Dogs}
        \label{fig:dogs_ratio}
    \end{subfigure}
    \begin{subfigure}{0.3\linewidth}
        \includegraphics[width=\linewidth]{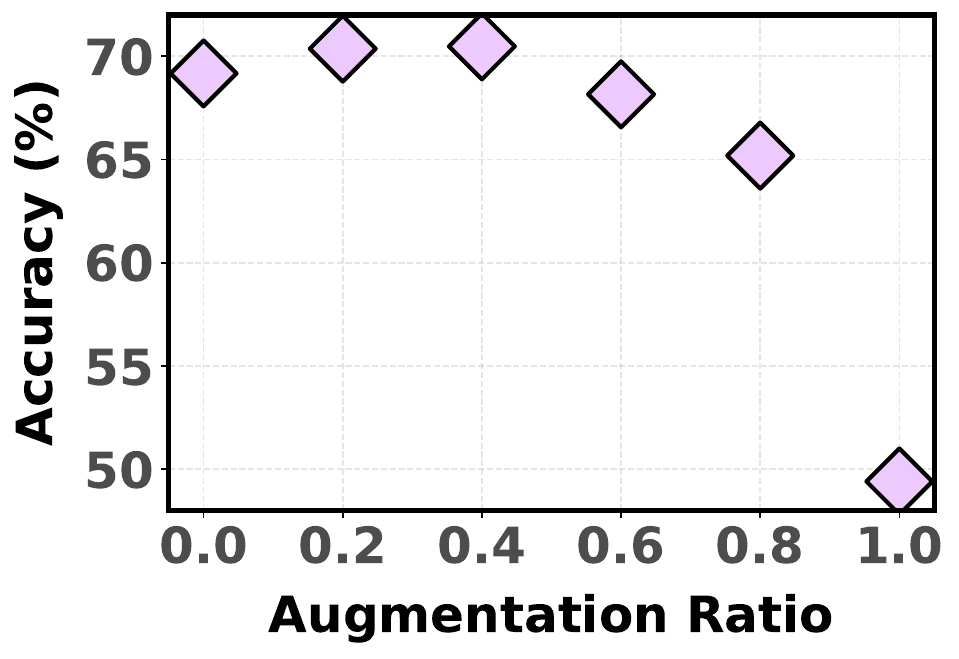}
        \caption{DTD}
        \label{fig:dtd_ratio}
    \end{subfigure}
    \caption{
    Impact of varying the augmentation ratio on accuracy across six datasets. 
    The ratio represents the proportion of synthetic images generated by \texttt{HiGFA} within the combined training set. 
    Accuracy is measured using the CAL classifier.
    }
    \label{fig:augmentation_ratio}
\end{figure*}

\Cref{fig:augmentation_ratio} presents the classification accuracy across the six datasets as a function of the augmentation ratio, defined as the proportion of synthetic images in the training set. 
Generally, peak performance is achieved when the synthetic data constitutes between $20\%$ and $60\%$ of the total training data. 
Accuracy tends to decline beyond this range. 
Notably, training exclusively on synthetic data results in a significant drop in performance compared to the optimal ratios for most datasets.

These findings suggest that while \texttt{HiGFA} generates high-quality and informative synthetic images, maintaining an appropriate balance between real and generated data is crucial. 
Moderate amounts of synthetic data can enhance the original dataset by increasing diversity and improving model generalization. 
However, an over-reliance on synthetic data, especially to the exclusion of real images, can be detrimental. 
The observed performance degradation at high augmentation ratios may result from subtle distributional shifts between the synthetic and real data or from overfitting to minor artifacts in the generated images, which become more prominent when real data is limited or unavailable.

\subsubsection{Performance on ViT-Based Fine-grained Visual Classifiers.}
\label{app:transfg}
To assess the generalizability of \texttt{HiGFA}, we evaluated its performance in augmenting data for a Vision Transformer (ViT)-based fine-grained classifier. 
We replaced the ResNet50-based CAL classifier used in the main experiments
with TransFG~\cite{he2022transfg}, a ViT method adapted for FGVC. 
All other experimental settings, including augmentation ratios and competitor implementations, were kept the same as CNN-Based experiments.

\begin{table*}[h]
\centering
\setlength{\tabcolsep}{3.0mm}{
\begin{tabular}{rlcccccc}
\toprule
\multicolumn{1}{c}{\multirow{2}[4]{*}{\textbf{Type}}} & \multicolumn{1}{l}{\multirow{2}[4]{*}{\textbf{Methods}}} & \multicolumn{3}{c}{{\textbf{Rigid Objects}}} & \multicolumn{3}{c}{{\textbf{Non-Rigid Objects}}} \\
\cmidrule{3-8}          &       & \textbf{Aircraft} & \textbf{Cars}  & \textbf{CompCars} & \textbf{CUB}   & \textbf{Dogs}  & \textbf{DTD} \\
\midrule
\multicolumn{1}{r}{\multirow{6}[2]{*}{Traditional}} 
& NoAug~\cite{he2022transfg}             & \orangee{79.8}  & \orangee{85.4}  & \orangef{56.8}  & \blued{87.9}  & \bluef{87.1}  & \bluef{71.5}  \\
& FRAug~\cite{he2022transfg}             & \orangee{80.2}  & \oranged{86.0}  & \orangee{58.4}  & \blued{87.9}  & \blued{87.7}  & \bluee{72.3}  \\
& CALAug~\cite{rao2021counterfactual}            & \orangef{79.2}  & \orangee{85.9}  & \oranged{60.1}  & \bluea{88.5}  & \blued{87.7}  & \blueb{73.9}  \\
& RandAug~\cite{cubuk2020randaugment}           & \orangec{81.5}  & \orangeb{\underline{87.2}}  & \orangeb{\underline{64.0}}  & \blueb{\textbf{88.8}}  & \blueb{87.8}  & \bluea{\textbf{74.5}}  \\
& AutoAug~\cite{autoaug}           & \orangeb{\underline{82.6}}  & \orangec{86.7}  & \orangec{62.9}  & \blueb{\underline{88.7}}  & \blueb{\underline{88.2}}  & \blueb{74.2}  \\
& RandEarse~\cite{zhong2020random}         & \oranged{80.8}  & \oranged{86.0}  & \orangee{58.6}  & \bluef{87.4}  & \bluec{88.1}  & \bluec{73.6}  \\
& CutMix~\cite{yun2019cutmix}            & \orangee{80.3}  & \orangec{86.2}  & \orangee{58.7}  & \bluec{88.1}  & \bluec{87.9}  & \blued{72.6}  \\
\midrule
\multicolumn{1}{r}{\multirow{6}[2]{*}{Generative}} 
& Real Guidance~\cite{he2022synthetic}     & \orangee{79.8}  & \orangeb{86.8}  & \orangeb{64.1}  & \bluec{88.2}  & \blueb{\underline{88.2}}  & \blueb{\underline{74.3}}  \\
& ALIA~\cite{dunlap2023diversify}             & \oranged{80.7}  & \orangef{83.9}  & \orangec{62.7}  & \bluee{87.6}  & \blueb{88.1}  & \blued{72.6}  \\
& DiffuseMix~\cite{islam2024diffusemix}        & \orangef{78.7}  & \orangee{84.8}  & \orangef{58.0}  & \bluee{87.8}  & \bluef{87.1}  & \blued{73.1}  \\
& DistDiff~\cite{yangdistribution}          & \orangec{81.8}  & \orangef{84.2}  & \oranged{60.2}  & \blueb{88.3}  & \bluee{87.4}  & \blueb{74.2}  \\
& SaSPA~\cite{michaeli2024advancing}             & \orangeb{82.5}  & \orangef{85.3}  & \orangeb{\underline{64.0}}  & \bluec{88.1}  & \bluec{87.9}  & \bluec{73.6}  \\
& \textbf{Ours}     & \orangea{\textbf{84.3}}  & \orangea{\textbf{87.6}}  & \orangea{\textbf{64.8}}  & \blueb{88.3}  & \bluea{\textbf{88.3}}  & \bluea{\textbf{74.5}}  \\
\bottomrule
\end{tabular}%
}
\caption{    
    Performance comparison of data augmentation methods using the TransFG (ViT-B) classifier across rigid and non-rigid datasets. 
    Results are reported as top-1 Accuracy (\%). 
    Best results are in \textbf{bold}, runner-up \underline{underlined}, respectively.
    }
\label{tab:fgvc_results_transfg}

\end{table*}

The accuracy results with the TransFG backbone are reported in \Cref{tab:fgvc_results_transfg}. 
\texttt{HiGFA} achieves the highest accuracy on five out of six benchmark datasets (Aircraft, Cars, CompCars, Dogs, and DTD). 
On the CUB dataset, it performs competitively, slightly below the best traditional augmentation method but outperforming other generative methods. 
These findings largely corroborate the trends observed with the ResNet backbone, indicating that the benefits of \texttt{HiGFA} are robust across different classifier architectures.

\subsubsection{Impact of Guidance Classifier Architecture.}
We investigated whether the architecture of the classifier providing fine-grained classifier guidance ($p_{\phi}$ in Eq. \ref{eq:classifier_guidance_ours}) impacts the final performance of the downstream classifier. 
We repeated the main experiments, replacing the ResNet101 guidance classifier with a Vision Transformer Base (ViT-B/32) model.
We used publicly available ViT models, fine-tuned for each dataset, from HuggingFace.
For CompCars, where no suitable public ViT model was found, we used a TransFG model trained with CALAug as the guidance classifier. 
The downstream classifier remained ResNet50-based CAL.

\begin{figure*}[h]
  \centering
  \includegraphics[width=0.75\linewidth]{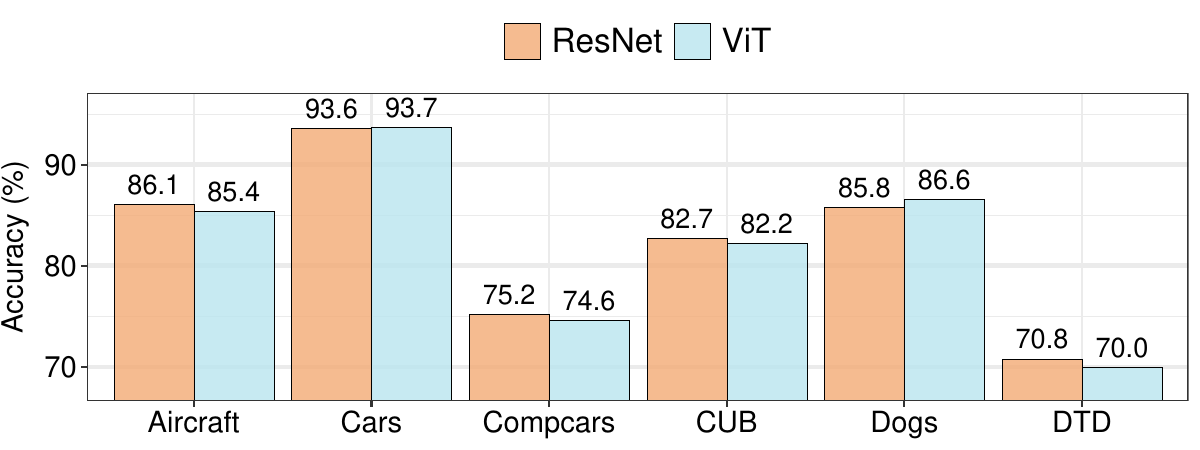} 
  \caption{
    Comparison of Accuracy using the CAL classifier when \texttt{HiGFA} employs either a ResNet-50 or a ViT-B model for its internal feature guidance. The results show high similarity, indicating robustness to the choice of guidance architecture.
  }
  \label{fig:ViTGuidance}
\end{figure*}

\Cref{fig:ViTGuidance} compares the downstream classification accuracy when using ResNet101 versus ViT-B for feature guidance. 
The results show a high degree of consistency: the final classification performance remains nearly identical regardless of whether the feature guidance in \texttt{HiGFA} is provided by ResNet or ViT. 
The observed performance differences across datasets are minimal, typically within $\pm 0.8\%$.
This suggests that \texttt{HiGFA} is robust to the choice of guidance  classifier architecture. 
As long as the classifier provides useful gradients, it can guide the generation toward the correct fine-grained category. 
The hierarchical and adaptive modules in \texttt{HiGFA} make good use of the guidance, whether from a convolutional or transformer-based model.

\subsection{Visualizations}
\label{app:visualizations}

\subsubsection{Qualitative Results on Transformed Contour Images.}
\begin{figure*}[h]
  \centering
  \begin{tikzpicture}
    \node[anchor=south west, inner sep=0] (main) at (0,0) {\includegraphics[width=0.95\linewidth]{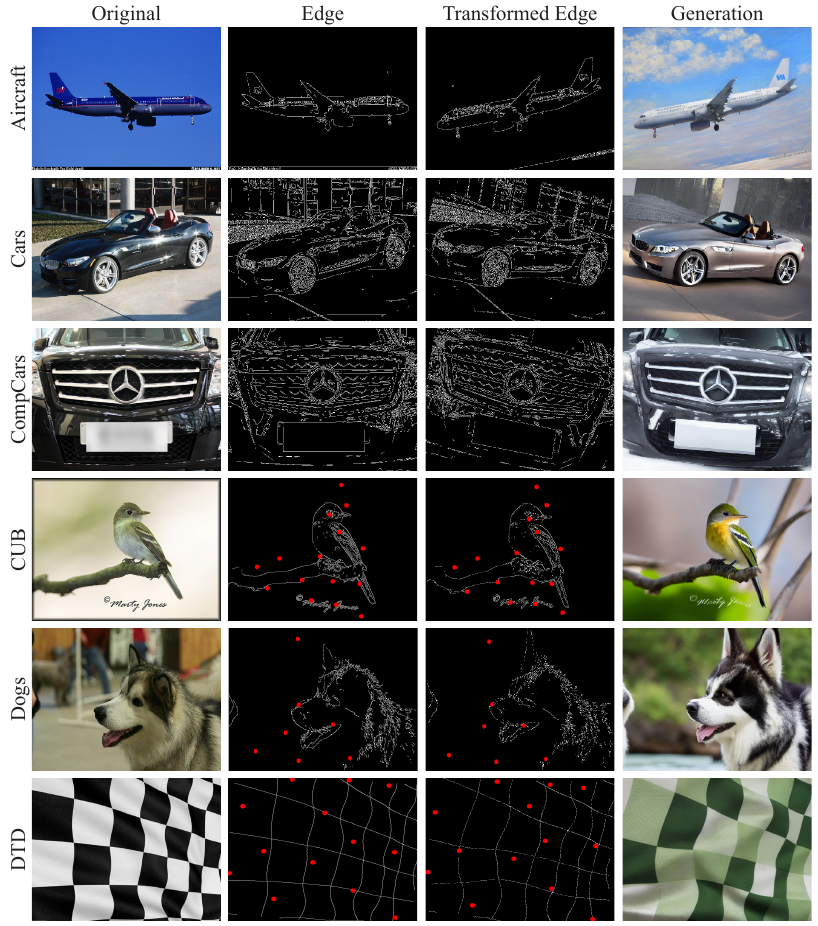}};
  \end{tikzpicture}
\caption{
Examples of transformed contour images are shown.
Random horizontal flipping and rotation modify object pose. 
The control points used for thin-plate spline interpolation are marked in the figure.
By applying thin-plate spline interpolation, the shape can be modified.
Transformations applied to contour maps help increase data diversity.
}
\label{fig:main_example}
\end{figure*}

\Cref{fig:main_example} illustrates the qualitative results of the transformed contour maps and the corresponding generated images. 
Random horizontal flipping and rotation are employed to effectively diversify object poses.
Additionally, applying thin-plate spline interpolation to the contours of non-rigid objects simulates realistic motion.
For instance, a bird’s head may appear flattened, a dog’s ears may become erect, or the folds in a cloth may shift.
These contour transformations significantly enhance the diversity and realism of the generated images.

\subsubsection{Qualitative Comparison with Different Guidance.}
\begin{figure*}[h]
  \centering
  \includegraphics[width=\linewidth]{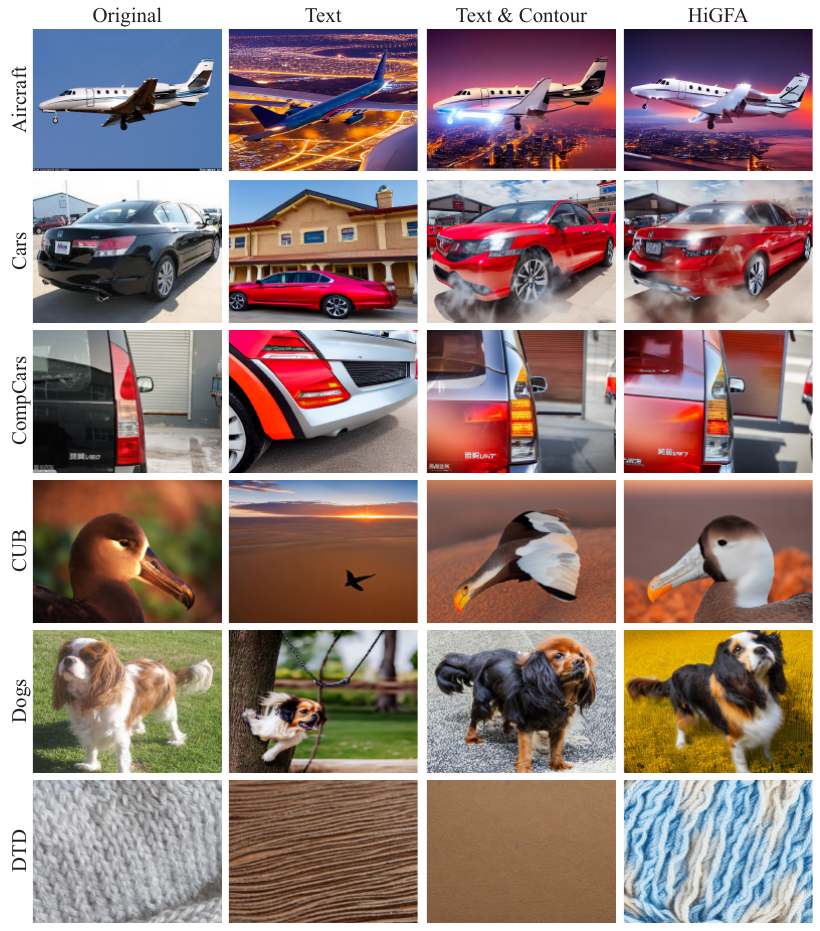} 
  \caption{
  Qualitative results with different guidance. 
  \textbf{a)} Using text guidance alone is insufficient to ensure category fidelity in the generated images. 
  \textbf{b)} Incorporating contour guidance improves category fidelity, but some subtle details remain inaccurate. 
  For example, missing regions near airplane wings, confusion between the front and rear of cars, or malformed bird heads. 
  \textbf{c)} Our method, \texttt{HiGFA}, which integrates all three types of guidance, further enhances both category fidelity and structural coherence in the generated results.
  }
  \label{fig:diff_guidance_vis}
\end{figure*}

Visualizations comparing images generated by \texttt{HiGFA} with different guidance components enabled as shown in \Cref{fig:diff_guidance_vis}, 
illustrate how each guidance type contributes to the final image quality and fidelity.
Relying solely on text guidance often results in low fidelity. 
Although contour guidance can enhance alignment, it may also lead to content omissions. 
Examples include missing regions near an airplane’s wing, confusion between the front and rear of a car, or distorted features such as a collapsed head in generated birds, possibly due to the model’s attempt to render a complete bird.

\subsubsection{Qualitative Comparison with Other Methods.}
\begin{figure*}[h]
  \centering
  \includegraphics[width=\linewidth]{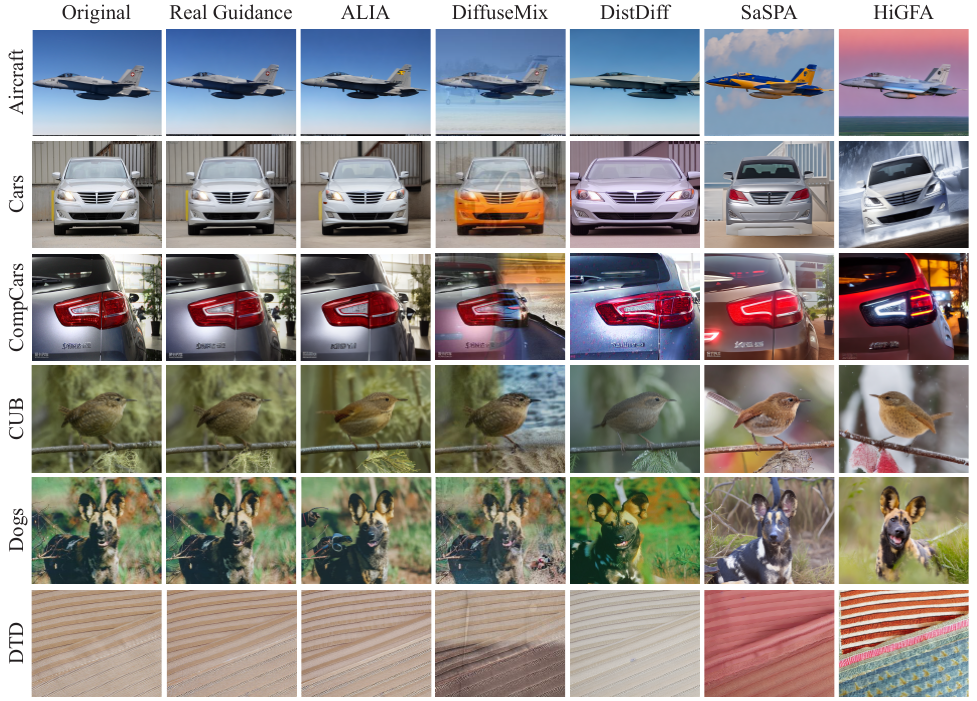} 
  \caption{
  Qualitative results of the original image and various generative augmentation methods, Real-Guidance, ALIA, DiffuseMix, DistDiff, SaSPA and \texttt{HiGFA} on six FGVC datasets are presented.
  \texttt{HiGFA} generates more realistic images with higher fidelity and greater diversity.
  }
  \label{fig:compair_vis}
\end{figure*}

Qualitative comparisons between images generated by \texttt{HiGFA} and baseline generative augmentation methods for fine-grained categories are shown in \Cref{fig:compair_vis}. 
These comparisons highlight \texttt{HiGFA}'s ability to generate diverse images while also better preserving subtle category-defining details.

\subsection{Limitations}
\label{sec:limitations}
Despite its strong performance, \texttt{HiGFA} has certain limitations.
First, although the multi-stage guidance mechanism is effective in enhancing the quality of augmented samples, it depends on the integration of multiple guidance components during the diffusion sampling process. 
While this contributes to improved robustness and sample fidelity, it also introduces architectural complexity, potentially limiting implementation flexibility compared to simpler single guidance methods.
Second, while the adaptive strategy improves robustness, the initial configuration of guidance strengths and the transition timestep for activating feature guidance may require empirical tuning to achieve optimal performance, particularly when adapting \texttt{HiGFA} to new datasets or to significantly different backbone diffusion models.
Finally, although well-trained fine-grained classifiers are publicly available on platforms such as HuggingFace and GitHub, applying the method to a new, private dataset necessitates training a fine-grained classifier beforehand, which adds further computational overhead.

\end{document}